\documentclass[letterpaper]{article} 
\usepackage{aaai2026}  
\usepackage{times}  
\usepackage{helvet}  
\usepackage{courier}  
\usepackage[hyphens]{url}  
\usepackage{graphicx} 
\usepackage{natbib}  
\usepackage{caption} 
\usepackage{algorithm}
\usepackage{algorithmic}

\usepackage{amsmath}
\usepackage{multirow}
\usepackage{makecell}
\usepackage{booktabs} 
\usepackage[table]{xcolor}
\usepackage{amssymb}
\usepackage{fontawesome}

\newcommand{\methodname}{LoViC}

\usepackage{newfloat}
\usepackage{listings}
\DeclareCaptionStyle{ruled}{labelfont=normalfont,labelsep=colon,strut=off} 
\floatstyle{ruled}
\newfloat{listing}{tb}{lst}{}
\floatname{listing}{Listing}
\title{\methodname: Efficient Long Video Generation with Context Compression}
\author{
	Jiaxiu Jiang\textsuperscript{\rm 1}, Wenbo Li\textsuperscript{\rm 2}\textsuperscript{\faStarO}, Jingjing Ren\textsuperscript{\rm 3}, Yuping Qiu\textsuperscript{\rm 3}, Yong Guo\textsuperscript{\rm 4}\\Xiaogang Xu\textsuperscript{\rm 2}, Han Wu\textsuperscript{\rm 5}, Wangmeng Zuo\textsuperscript{\rm 1}\textsuperscript{\faEnvelopeO}\\
}
\affiliations{
\textsuperscript{\rm 1}Harbin Institute of Technology$\quad$
\textsuperscript{\rm 2}The Chinese University of Hong Kong$\quad$
\textsuperscript{\rm 4}Max Planck Institutes\\
\textsuperscript{\rm 3}The Hong Kong University of Science and Technology (Guangzhou)$\quad$
\textsuperscript{\rm 5}Independent Researcher\\
\textsuperscript{\faStarO}Project Leader$\quad$\textsuperscript{\faEnvelopeO}Corresponding Author\\
\texttt{Project Page: https://jiangjiaxiu.github.io/lovic/}
}

\begin{document}

\twocolumn[{
	\renewcommand\twocolumn[1][]{#1}
	\maketitle
	\begin{center}
		\centering
		\captionsetup{type=figure}
		\includegraphics[width=0.95\linewidth]{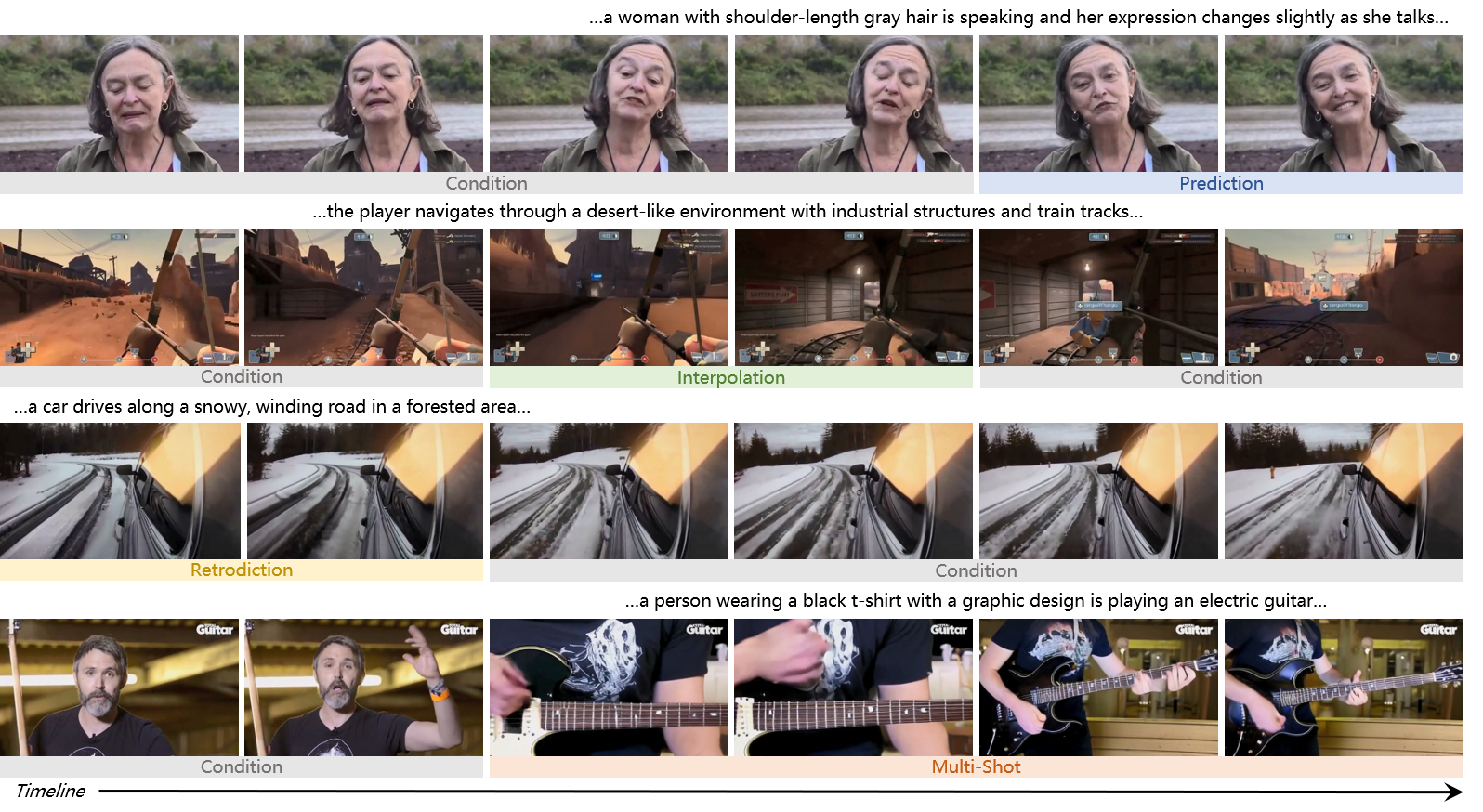}
		\captionof{figure}{
			\textbf{Videos generated by our model.}
			Our model has the flexibility to do video continuation of any direction and generate multi-shot video with advanced efficiency. It shows the capability of retaining ID consistency within large temporal range, generating videos of large and smooth motion.
		}
		\label{fig:teaser}
	\end{center}
}]

\begin{abstract}
Despite recent advances in diffusion transformers (DiTs) for text-to-video generation, scaling to long-duration content remains challenging due to the quadratic complexity of self-attention. While prior efforts---such as sparse attention and temporally autoregressive models---offer partial relief, they often compromise temporal coherence or scalability. We introduce \methodname, a DiT-based framework trained on million-scale open-domain videos, designed to produce long, coherent videos through a segment-wise generation process. At the core of our approach is FlexFormer, an expressive autoencoder that jointly compresses video and text into unified latent representations. It supports variable-length inputs with linearly adjustable compression rates, enabled by a single query token design based on the Q-Former architecture. Additionally, by encoding temporal context through position-aware mechanisms, our model seamlessly supports prediction, retradiction, interpolation, and multi-shot generation within a unified paradigm. Extensive experiments across diverse tasks validate the effectiveness and versatility of our approach.
\end{abstract}

\section{Introduction}

Recent advances in text-to-video (T2V) generation have demonstrated impressive capabilities in synthesizing short video clips~\cite{polyak2024movie, wang2025wan, zheng2024open, ma2025step, liu2025lumina, hacohen2024ltx, lin2024open, kong2024hunyuanvideo, yang2024cogvideox, zhou2024allegro, jin2024pyramidal}, which leverage large-scale video-text datasets and scalable diffusion transformer (DiT) architectures~\cite{peebles2023scalable, bao2023all}. However, many real-world applications---such as first-person driving simulations, immersive 3D games, product promotional content, and user-generated storytelling---demand coherent long-form video generation. Naively extending DiT models to longer sequences is computationally prohibitive due to their quadratic self-attention cost and the inherent density of tokenized video representations, while alternative ad-hoc strategies often compromise temporal coherence and content consistency.

\begin{figure}
	\centering
	\includegraphics[width=0.95\linewidth]{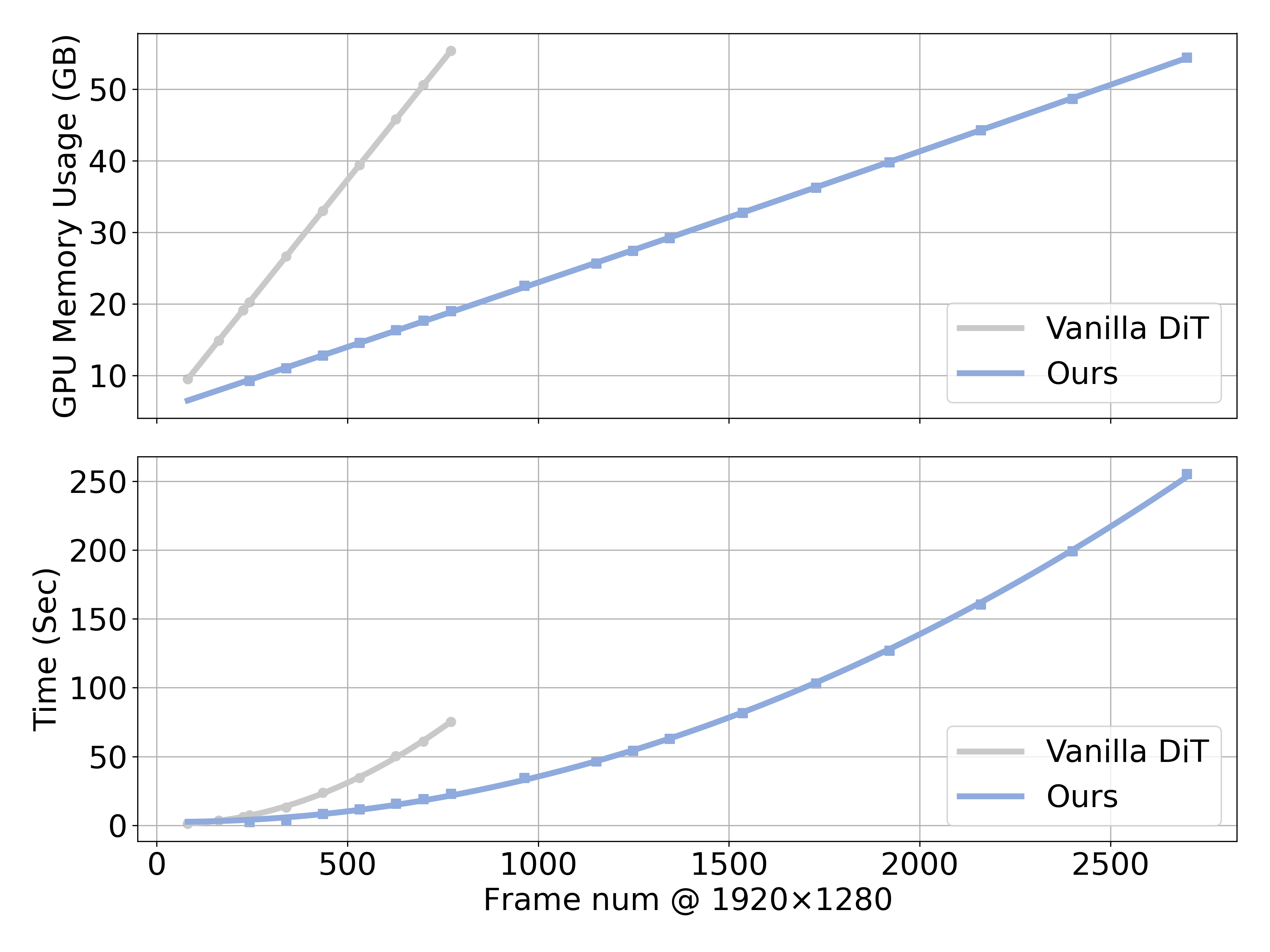}
	\caption{\textbf{Memory and time usage of single timestep DiT inference.}
	With context compression, our method reduces memory usage and runtime, allowing more frames to be generated within the same resource constraints.
	}
	\label{fig:memory}
\end{figure}

A common strategy for long video generation is to synthesize short clips in a segment-by-segment manner, where the key challenge lies in efficiently scaling the context length---defined as the temporal span of the conditioning video that determines the model’s receptive field. Crucially, the appropriate context length is task-dependent and varies with the temporal complexity and dynamics of the scene.

In relatively simple scenarios, such as natural scene videos with low temporal complexity, segments can often be generated independently and stitched together with minimal loss of coherence~\cite{zhu2023moviefactory}. For moderately complex cases---\textit{e.g.}, first-person driving videos or Minecraft gameplay streams---a block-autoregressive approach~\cite{voleti2022mcvd, harvey2022flexible} can yield decent results, provided that transitions between blocks are smooth. In these cases, short-range context is typically sufficient to maintain temporal consistency.

However, in more challenging settings with high visual and temporal complexity, maintaining content consistency often requires long-range temporal dependencies---\textit{i.e.}, a context window spanning more than two segments~\cite{henschel2024streamingt2v}. To address this, several methods have attempted to extend the effective context. For instance, FDM~\cite{harvey2022flexible} enhances block-wise autoregressive generation by sampling frames scattered across a broader temporal range in each step. NUWA-XL~\cite{yin2023nuwa} adopts a two-stage strategy by first generating sparse keyframes and then interpolating them in a coarse-to-fine manner. Despite these efforts, existing approaches have not demonstrated strong performance on open-domain, high-variability video generation tasks. These scenarios often demand substantially longer context windows per generation step to preserve both semantic and temporal consistency across frames.

In this work, we propose a general framework, \methodname, for long-form video generation that accommodates a wide range of video settings. At each generation step, our model conditions on the full history of preceding video segments to synthesize the next segment. To mitigate the computational burden of self-attention over long temporal contexts, we introduce FlexFormer, a flexible Q-Former-based encoder that compresses context of arbitrary length---including both video and text tokens---under an adaptive compression ratio. FlexFormer employs a single learnable query token and a novel Interpolated RoPE (I-RoPE) positional encoding scheme to efficiently summarize long-range information. The resulting compressed context features are fed into a DiT-based decoder to generate the current video chunk. Our unified architecture supports a broad spectrum of generation tasks, including video prediction, retrodiction, interpolation, and multi-shot generation, by framing them under a single, coherent paradigm. We train our model on a million-scale open-domain video corpus and demonstrate its effectiveness and generality by extensive experiments.

Our contributions are as follows:
\begin{itemize}
	
	\item We propose a novel framework, LoViC, that supports extended temporal contexts in DiTs-based video generation, enabling more coherent long-form video synthesis.
	
	\item To efficiently encode long video-text context, we introduce FlexFormer, a flexible autoencoder that supports inputs of arbitrary shape with adaptive compression rates. FlexFormer employs a single-query token mechanism alongside a novel I-RoPE positional encoding scheme, which preserves spatial relationships among video tokens during compression.
	
	\item Our framework supports both single-shot video extension and multi-shot video generation by introducing relative positional encoding to distinguish contextual segments from the current segment, enabling visually coherent results and realistic dynamics across diverse tasks.
	
\end{itemize}

\section{Related Work}

\paragraph{Video Diffusion Models.}
Earlier approaches~\cite{ho2022video, ho2022imagen, bar2024lumiere, blattmann2023align, guo2023animatediff, wang2023modelscope, wang2025lavie, chen2023videocrafter1, blattmann2023stable, chen2024videocrafter2} adopt a U-Net~\cite{ronneberger2015u} architecture to model the diffusion process in the latent space of a VAE~\cite{rombach2022high} for video generation.
More recent methods~\cite{polyak2024movie, wang2025wan, zheng2024open, ma2025step, liu2025lumina, hacohen2024ltx, lin2024open, kong2024hunyuanvideo, yang2024cogvideox, zhou2024allegro, jin2024pyramidal} replace the U-Net with transformer~\cite{peebles2023scalable, bao2023all} to improve scalability.

\begin{figure*}[tb]
	\centering
	\includegraphics[width=0.9\linewidth]{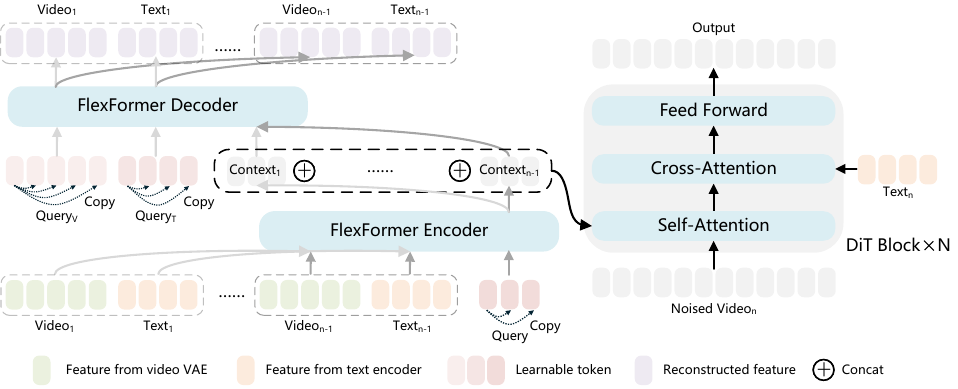}
	\caption{\textbf{Model architecture.}
		The left part of the figure features an autoencoder consisting of a FlexFormer encoder and a FlexFormer decoder. The encoder compresses multiple segments of video and text tokens separately. The number of query tokens is derived from the video token sequence length. The query token sequence is formed by copying the single learnable token multiple times. The decoder decode the context tokens into video and text features in a similar way. Each context video-text pair is compressed into some context tokens. The multiple chunks of context tokens are concatenated then fed into the DiT by further concatenating with the input tokens of the self-attention layer.
	}
	\label{fig:model}
\end{figure*}

\paragraph{Long Video Generation.}
We focus on long video generation using diffusion models. Previous work has proposed roughly two categories of approaches to mitigate quadratic complexity of the vanilla DiT.
The first category adopts a divide-and-conquer strategy, dividing the frames of a long video into several portions and generate a portion each time. Most of these methods generate long videos autoregressively by concatenating conditioning frames and current frames~\cite{harvey2022flexible, yin2023nuwa, chen2023seine, jin2024pyramidal, gao2024vid, li2024arlon, zhuang2025video, yu2025malt, song2025history, xiao2025worldmem, yin2024slow, xie2024progressive, chen2025skyreels, zhang2025packing, hu2024acdit, gu2025long, zhang2025packing, ruherolling, gao2024ca2, zhang2025generative, teng2025magi, huang2025self, lin2025autoregressive}. Frame-causal or block-causal attention mechanisms are typically applied in this setting. Depending on how the conditioning and current frames are arranged, these models can perform either video prediction or interpolation. Among them, Pyramidal Flow~\cite{jin2024pyramidal} and FramePack~\cite{zhang2025packing} additionally explore compressing historical frames to improve efficiency.
Another line of work encodes historical context with auxiliary modules~\cite{voleti2022mcvd, henschel2024streamingt2v, xiang2024pandora, ouyang2024flexifilm, sun2024video, savov2025statespacediffuser0}. Some of these models, such as Pandora~\cite{xiang2024pandora} and StateSpaceDiffuser~\cite{savov2025statespacediffuser0}, also concatenate a few adjacent conditioning frames to enhance temporal continuity.

The second category of methods replaces self-attention with more efficient architectures. Some methods adopt hybrid designs that combine local self-attention with linear attention variants~\cite{wang2025lingen, dalal2025one, zhang2025test}, while others employ alternative mechanisms such as overlapped window attention~\cite{zheng2025frame} or sparse attention~\cite{zhang2025training}.

\paragraph{Multi-Shot Video Generation.}
A parallel research direction focuses on generating a specific form of long video---multi-shot videos. Most methods leverage temporally-dense captions to generate multiple shots within the original video length of the pretrained DiT~\cite{bansal2024talc, kara2025shotadapter, qi2025mask, wang2025echoshot}. LCT~\cite{wu2024mind} takes a different approach by directly training vanilla DiT on longer videos with abundant resources. TTT~\cite{dalal2025one} and Frame-Level Captions~\cite{zheng2025frame} adopts efficient attention architectures to support longer videos. Another group of methods~\cite{zhou2024storydiffusion, zhuang2024vlogger, long2024videostudio, zheng2024videogen, xiao2025videoauteur} follows a two-stage pipeline: first generating a sequence of coherent keyframes, and then applying image-to-video generation to produce the final video.

Additionally, training-free methods~\cite{ho2022video, oh2024mevg, kim2025tuning, kang2025text2story, gu2023reuse, wang2023gen, wang2024zola, qiu2023freenoise, lufreelong, li2024vstar, kim2025fifo, chen2025ouroboros, zhao2025riflex, fang2025inflvg} have been proposed for generating long and multi-shot videos.

\section{Method}

\begin{figure*}
	\centering
	\includegraphics[width=0.95\linewidth]{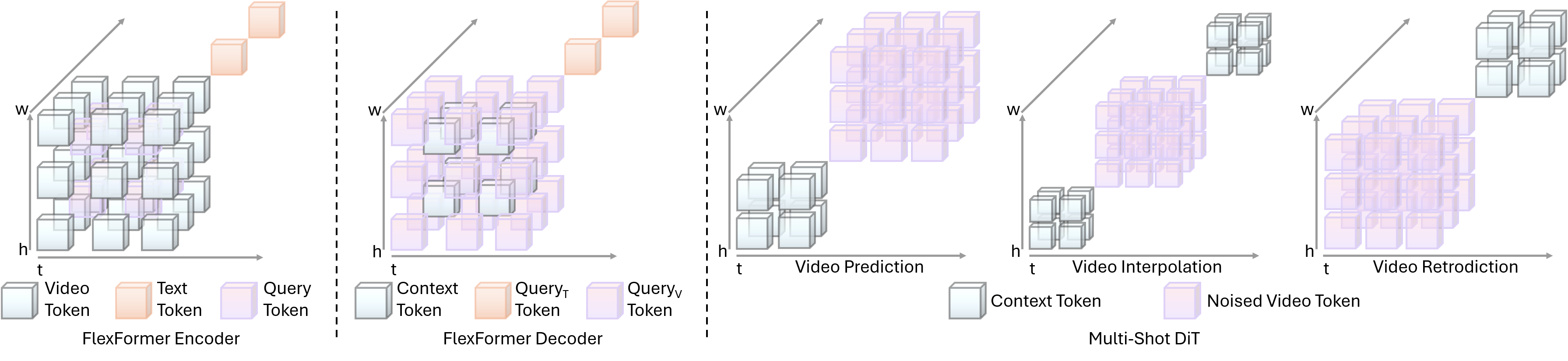}
	\caption{\textbf{Illustration of our positional encoding.}
		Each block represents the positional index $({t}, {h}, {w})$ of the corresponding token. The illustrated compression strategy is uniform compression. To adapt for multi-shot generation, the blue and purple blocks will be separated slightly.
	}
	\label{fig:rope}
\end{figure*}

\subsection{Preliminaries}

\subsubsection{Diffusion Transformer.}
Diffusion models with transformers (DiTs)\cite{peebles2023scalable} have emerged as a powerful architecture that integrates the strengths of diffusion probabilistic models with transformer-based representations. DiTs are commonly trained using the Flow Matching objective\cite{esser2024scaling, liu2022flow, albergo2022building, lipman2022flow}:
\begin{gather}
	\mathcal{L}_{flow} = ||(\epsilon-z_0)-u_{\theta}(z_t,t,c)||^2 \,, \\
	z_t = t\cdot z_0 + (1-t)\cdot \epsilon \,,
\end{gather}
where \( u_{\theta} \) denotes the DiT model, \( \epsilon \sim \mathcal{N}(0, I) \) is Gaussian noise, \( z_0 \) is a data sample, \( t \in [0, 1] \) is the noise interpolation timestep, and \( c \) represents a conditioning signal (\textit{e.g.}, text prompts). In text-to-video applications, DiT models typically encode text in two ways: via cross-attention mechanisms~\cite{polyak2024movie, hacohen2024ltx, wang2025wan} or by integrating text directly through self-attention~\cite{kong2024hunyuanvideo, yang2024cogvideox}. Regardless of the modality interaction strategy, the quadratic complexity of vanilla attention over dense spatiotempora tokens remains a key bottleneck, limiting the scalability of DiT models for long-form video generation.

\paragraph{M-RoPE.}
M-RoPE~\cite{bai2025qwen2} extends 3D-RoPE~\cite{su2024roformer} to handle multi-modal token sequences by unifying 1D (text) and 2D (image) data as special cases of 3D positional encoding. In this setting, an image is treated as a single-frame video, and a text token is seen as a single-pixel video frame, effectively treating them as pseudo-3D tokens. Similar positional strategies have also been adopted in multi-shot video generation~\cite{guo2025long}. While M-RoPE is conceptually versatile, we empirically demonstrate that it performs suboptimally under certain conditions.

Given a token embedding $\mathbf{x}\in \mathbb{R}^{d}$ and its position index $p \in \mathbb{N}$, RoPE transforms $\mathbf{x}$ into:
\begin{equation}
	\text{RoPE}(\mathbf{x}, p) = \mathbf{x}_{\text{even}} \circ \cos(\boldsymbol{\theta}(p)) + \mathbf{x}_{\text{odd}} \circ \sin(\boldsymbol{\theta}(p)) \,,
\end{equation}
where $\mathbf{x}_{\text{even}}$ and $\mathbf{x}_{\text{odd}}$ are even and odd dimensions of $\mathbf{x}$, respectively, and
$\boldsymbol{\theta}_i(p) = \frac{p}{10000^{2i/d}}, i = 0, 1, \ldots, \frac{d}{2} - 1$. 3D-RoPE decomposes token's spatiotemporal position $\mathbf{p}$ into three components: temporal $p_t$, height $p_h$, width $p_w$. Token embedding $\mathbf{x}$ is divided into three equal subvectors:
\begin{equation}
	\mathbf{x} = \mathbf{x}^{(t)} \oplus \mathbf{x}^{(h)} \oplus \mathbf{x}^{(w)}, \quad \mathbf{x}^{(\cdot)} \in \mathbb{R}^{d/3} \,,
\end{equation}
where $\oplus$ denotes the concatenation of vectors. Then we apply RoPE independently to each axis:
\begin{equation}
	\begin{split}
		&\text{M-RoPE}(\mathbf{x}, \mathbf{p}) = \text{3D-RoPE}(\mathbf{x}, \mathbf{p}) =\\
		&\text{RoPE}(\mathbf{x}^{(t)}, p_t) \oplus \text{RoPE}(\mathbf{x}^{(h)}, p_h) \oplus \text{RoPE}(\mathbf{x}^{(w)}, p_w) \,.
	\end{split}
\end{equation}

\subsection{Context Compression with FlexFormer}
When generating a new video segment, we condition on previously generated segments along with their corresponding text prompts. Prior work~\cite{guo2025long, wu2024mind, kara2025shotadapter, yan2024long, qi2025mask, dalal2025one} has demonstrated the importance of leveraging such context to ensure temporal and semantic consistency. To support arbitrarily ordered and variably sized context sequences, we design FlexFormer, a flexible autoencoder built upon the Q-Former architecture with several key enhancements. The vanilla Q-Former faces two primary limitations: (1) its cross-attention design prevents the application of RoPE, hindering spatial awareness across context tokens; and (2) its use of a fixed-length learnable query sequence can result in substantial information loss when encoding long video contexts. To address these limitations, we propose two architectural modifications that enable FlexFormer to compress multi-modal token sequences of arbitrary size and ordering, while adapting compression ratios dynamically.

\paragraph{Single Learnable Query Token.}
To support arbitrarily sized video-text context, we introduce a single learnable query token design. Instead of using cross-attention as in Q-Former, we adopt a self-attention-based encoding scheme inspired by MMDiT~\cite{esser2024scaling}. Specifically, the learnable query token is first replicated multiple times and concatenated with the video and text tokens. The number of copies is determined by the total number of input tokens and the specified compression strategy. To differentiate among the replicated query tokens, we apply positional embeddings---described in detail in the next part---to inject spatiotemporal cues after the encoding process. For decoding, we introduce two learnable query tokens (one for video and one for text), which are also replicated to match the original length of the respective token sequences. When handling multiple segments, \textit{e.g.}, $\{(\text{Video}_1,\text{Text}_1), \dots, (\text{Video}_{n-1}, \text{Text}_{n-1})\}$ as shown in Figure~\ref{fig:model}, we encode each pair independently and combine the compressed tokens $\{\text{Context}_1, \dots, \text{Context}_{n-1}\}$ before passing them to the DiT model.

\begin{figure}
	\centering
	\includegraphics[width=0.95\linewidth]{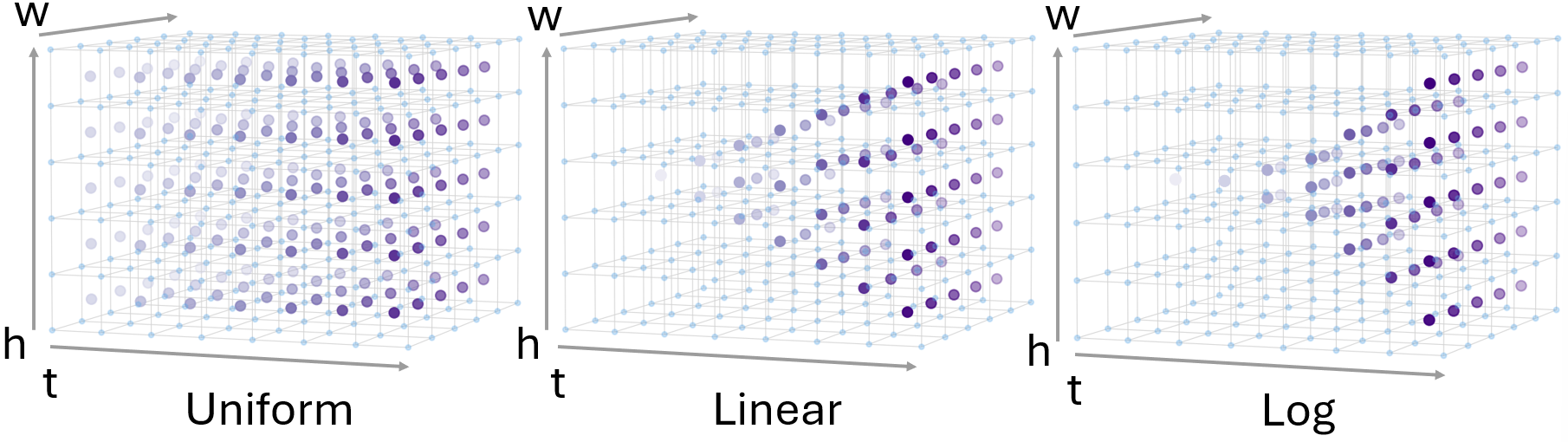}
	\caption{\textbf{Illustration of different compression strategies.} Blue and purple dots represent the position of video tokens and query tokens respectively. Text tokens are omitted.
	}
	\label{fig:compression_strategy}
\end{figure}

\paragraph{Interpolated-RoPE (I-RoPE).}
For multi-modal token sequences consisting of video, text, and query tokens, a common strategy for encoding positional relationships is M-RoPE~\cite{bai2025qwen2}, which has shown strong performance in MLLMs. However, when query tokens are treated as 1D tokens—analogous to text---M-RoPE performs suboptimally in our setting, often resulting in blurry reconstructions. We hypothesize that this degradation arises from the lack of an explicit spatial prior, making it difficult for the transformer to preserve the spatial structure of video content. To address this, we propose I-RoPE:
\begin{gather}
	\text{I-RoPE}(q_i) = \text{M-RoPE}(q_i, (p_t^{(i)}, p_h^{(i)}, p_w^{(i)})) \,,\\
	(p_t^{(i)}, p_h^{(i)}, p_w^{(i)}) = \text{Interpolate}(i; T, H, W, N) \,,
\end{gather}
which assigns spatiotemporal positions to query tokens by interpolating the indices of video tokens (see Figure~\ref{fig:rope}). The decoder employs a symmetric positional encoding scheme to that of the encoder.

As shown in Figure~\ref{fig:compression_strategy}, different interpolation styles enable different compression strategies. In the left panel, query tokens are uniformly distributed among video tokens. In the middle panel, the density of query tokens are linearly-scaled with respect to time. This design allocates lower compression ratios to frames adjacent to the current segment, which improves temporal continuity in the generated video, while maintaining high computational efficiency.

\subsection{Multi-Shot DiT}

We aim to adapt the vanilla DiT architecture to a broad range of video generation tasks with minimal architectural changes. Specifically, we decompose long-form video generation into four sub-tasks: video prediction, interpolation, retrodiction, and multi-shot video generation. Context features produced by the FlexFormer encoder are injected into the DiT via the self-attention layers of each transformer block. These features are concatenated with the original input tokens to the self-attention layers. This design choice offers two key advantages: (1) it facilitates better differentiation of the temporal positions of individual video clips, and (2) it avoids increasing model size, as no additional cross-attention layers are required.

As shown in the third section of Figure~\ref{fig:rope}, we assign different RoPE indices to the context tokens to distinguish among various generation tasks. For multi-shot video generation, we introduce a temporal gap between distinct shots---resulting in greater separation between the blue and purple blocks in the positional space.

\section{Experiments}

\begin{figure}[tb]
	\centering
	\includegraphics[width=\linewidth]{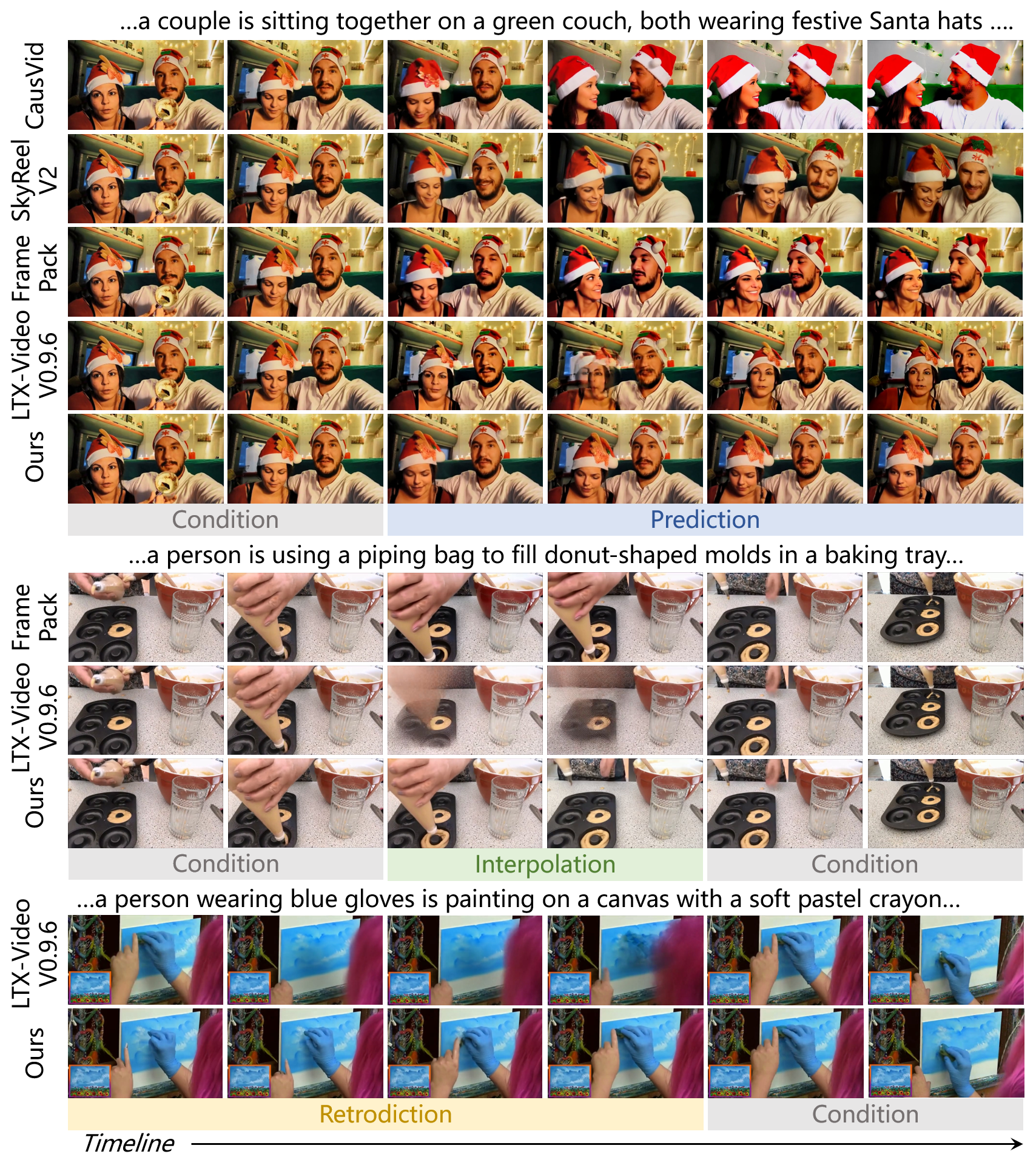}
	\caption{\textbf{Comparison on video prediction, interpolation and retrodiction.}
		Our model demonstrates the best consistency between conditioning video and generated video and produces a smooth transition between the videos.
	}
	\label{fig:single_shot}
\end{figure}

\subsection{Experimental Setup}

\paragraph{Dataset Construction.}
We utilize the open-source open-world video-text pair dataset Panda-70M~\cite{chen2024panda}, which builds the ASR-captioned video-text dataset HD-VILA-100M~\cite{xue2022advancing}. The original videos, sourced from YouTube, are predominantly single-shot with an average duration of 8.5 seconds. While Panda-70M provides captions generated with their proposed pipeline, we found these captions too short to guide our base model in generating high-quality videos. To obtain more informative and descriptive captions, we re-captioned the dataset using Qwen2.5-VL-7B~\cite{bai2025qwen2}. Panda-70M provides a subset of 2M video clips, from which we managed to download 1.6M video clips to train our model. To save resources, we retrieved videos at 360P resolution. Each group of three clips in the subset originates from the same long YouTube video, which makes it suitable for multi-shot video generation. However, as these clips are randomly sampled, the triplets do not always exhibit clear visual or semantic consistency. To address this, we employ ViCLIP~\cite{wang2023internvid} to filter out video pairs of cosine similarity below $0.5$, which are approximately half the dataset.

For video prediction, interpolation and retrodiction tasks, we further segment each single-shot video in the 2M subset into three equal-length clips, as the tasks require at least three temporal segments. Considering that our base model supports a maximum of 257 frames per generation, we clip each video to contain between 81 to 257 frames, resulting in a maximum total video length of 771 frames.

\begin{figure}[tb]
	\centering
	\includegraphics[width=\linewidth]{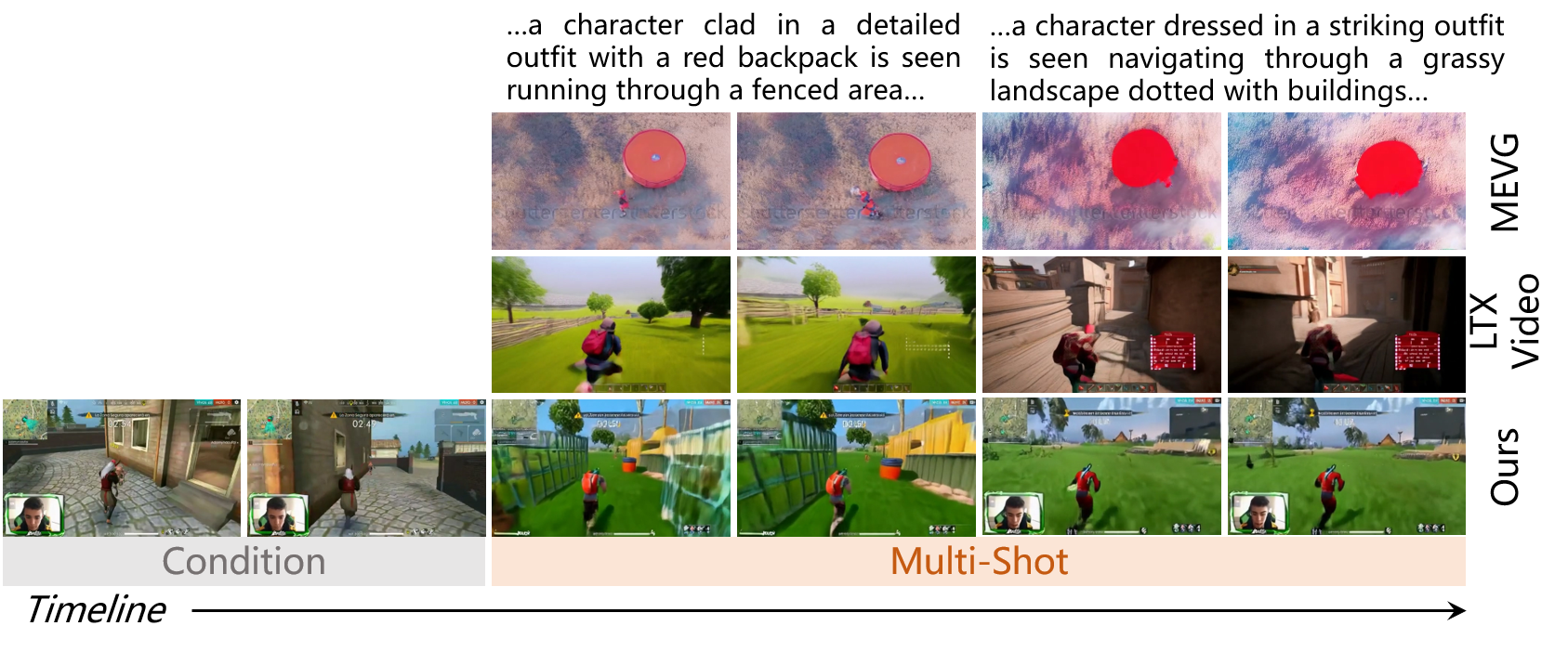}
	\caption{\textbf{Comparison on multi-shot video generation.}
		The baselines do not support conditioning on video.
	}
	\label{fig:multi_shot}
\end{figure}

\paragraph{Evaluation Settings.}
We evaluate the methods on the test set of Panda-70M.
For video prediction, interpolation and retrodiction, we randomly select 100 single-shot videos from the test set and clip each of them into three clips. For video prediction, we condition on the first two clips and generate the last clip. For interpolation, we condition on the first and last video, then generate the middle clip. For video retrodiction, we conditon on the last two clips and generate the first clip. 
We select two sets of metrics for evaluation. To evaluate whether the generated video clips seamlessly extend the given conditional video, we report PSNR, SSIM and LPIPS between the generated videos and the ground truth videos, following previous work~\cite{gu2025long, xiao2025worldmem}. To evaluate the general quality of the generated video, we select five metrics from VBench~\cite{huang2024vbench}, which are Subject Consistency, Background Consistency, Motion Smoothness, Aesthetic Quality and Video-text Alignment.

For multi-shot video generation, we random select 100 sets of prompts from the test set. We first generate the first video clip with our base model, then autoregressively generate the second and third clips. To evaluate the quality of the multi-shot videos, we report four metrics from the VBench, which are Subject Consistency, Background Consistency, Aesthetic Quality and Video-text Alignment, following previous work~\cite{dalal2025one, guo2025long}.

\begin{table*}
	\scriptsize
	\centering
	\begin{tabular}{c|c|cc|ccccc}
		\toprule
		\multirow{3}{*}{\makebox[0.084\textwidth]{\textbf{Method}}} & \multirow{3}{*}{\makebox[0.03\textwidth]{\makecell{\textbf{Context}\\ \textbf{Length}}}} &  \multicolumn{2}{c|}{\textbf{Reference-based Metrics}} & \multicolumn{5}{c}{\textbf{Non-reference Metrics}}\\
		~ & ~ & PSNR\textuparrow  & LPIPS\textdownarrow & \makecell{Subject\\Consistency\textuparrow} & \makecell{Background\\Consistency\textuparrow} & \makecell{Motion\\Smoothness\textuparrow} & \makecell{Aesthetic\\Quality\textuparrow} & \makecell{Video-text\\Alignment\textuparrow} \\
		\midrule
		\makebox[0.084\textwidth]{CausVid(1.4B)} & 16 & 11.31/--/-- & 0.491/--/-- & 0.909/--/-- & 0.916/--/-- & 0.986/--/-- & \textbf{0.508}/--/-- & \textbf{0.243}/--/-- \\
		\makebox[0.084\textwidth]{SkyReel-V2(1.4B)} & 17 & 14.81/--/-- & 0.376/--/-- & 0.920/--/-- & 0.941/--/-- & 0.991/--/-- & 0.476/--/-- & \underline{0.240}/--/-- \\
		\makebox[0.084\textwidth]{FramePack(13B)} & 19 & 14.40/15.63/-- & \underline{0.343}/0.308/-- & \underline{0.963}/\underline{0.934}/-- & \underline{0.966}/\underline{0.952}/-- & \underline{0.995}/\textbf{0.995}/-- & \underline{0.490}/\textbf{0.490}/-- & 0.238/\textbf{0.230}/-- \\
		\makebox[0.084\textwidth]{LTX-Video(1.9B)} & $\ast$ & \makebox[0.095\textwidth]{\underline{15.70}/\underline{16.77}/\underline{15.53}} & \makebox[0.095\textwidth]{0.344/\underline{0.302}/\underline{0.346}} & \makebox[0.095\textwidth]{0.897/0.896/\underline{0.920}} & \makebox[0.095\textwidth]{0.937/0.940/\underline{0.950}} & \makebox[0.095\textwidth]{0.989/0.982/\underline{0.990}} & \makebox[0.095\textwidth]{0.454/0.450/\underline{0.453}} & \makebox[0.095\textwidth]{0.220/0.213/\underline{0.214}} \\
		\rowcolor{gray!15}
		\makebox[0.084\textwidth]{Ours(2.3B)} & $\ast$ & \makebox[0.095\textwidth]{\textbf{15.76}/\textbf{16.82}/\textbf{15.60}} & \makebox[0.095\textwidth]{\textbf{0.316}/\textbf{0.272}/\textbf{0.325}} & \makebox[0.095\textwidth]{\textbf{0.981}/\textbf{0.947}/\textbf{0.985}} & \makebox[0.095\textwidth]{\textbf{0.970}/\textbf{0.954}/\textbf{0.974}} & \makebox[0.095\textwidth]{\textbf{0.996}/\textbf{0.995}/\textbf{0.996}} & \makebox[0.095\textwidth]{0.485/\underline{0.479}/\textbf{0.477}} & \makebox[0.095\textwidth]{0.237/\underline{0.228}/\textbf{0.220}} \\
		\bottomrule
	\end{tabular}
	\caption{\textbf{Comparison on video prediction/interpolation/retrodiction.} The best performance is \textbf{bold} and the second best is \underline{underlined}. The ''$-$'' means the method is not capable of the setting. Context length is measured by maximum frame number of conditioning video, and $\ast$ represents unfixed context length.}
	\label{tab:singleshot}
\end{table*}

\paragraph{Baselines.}
For video prediction, interpolation and retrodiction, we compare our method with CausVid~\cite{yin2024slow}, SkyReel-V2~\cite{chen2025skyreels}, LTX-Video-V0.9.6~\cite{hacohen2024ltx} and FramePack~\cite{zhang2025packing}.  CausVid finetunes the bidirectional self-attention in the text-to-video DiT Wan~\cite{wang2025wan} into causal-attention to enable block-autoregressive generation, where each block contains 12 frames. We take the last 12 frames of the conditioning video as the first block for CausVid to predict the subsequent video clip. SkyReel-V2 finetunes Wan with Diffusion Forcing~\cite{chen2024diffusion} to perform block-autoregressive generation similar to CausVid. It is able to condition on a maximum of 97 frames. LTX-Video-V0.9.6 is an updated version of LTX-Video that supports video extension by jointly denoising the conditioning video and the sampled noise in an MMDiT-like manner. It supports the longest context length among all baselines. Our experiments show that LTX-Video-V0.9.6 can utilize the entire conditioning video as input, though this results in higher computational cost, which limits its scalability for longer video generation. FramePack finetunes the HunyuanVideo~\cite{kong2024hunyuanvideo} into a block-autoregressive model with a reduced context window, aiming to lower the computational cost of the 13B parameter model. Except for FramePack, all other baselines use models with approximately 2B parameters.

For multi-shot video generation, we compare against two baselines following the setup in MinT~\cite{wu2024mind}. The first baseline uses our base model LTX-Video-V0.9.0~\cite{hacohen2024ltx} to generate the individual shots independently and then concatenates them into a single video. The second is MEVG~\cite{oh2024mevg}, a state-of-the-art training-free method that uses the DDIM inversion~\cite{song2020denoising} of the previously generated video clip as to initialize the noise for the next shot.

\begin{table}
	\scriptsize
	\centering
	\begin{tabular}{c|cccc}
		\toprule
		\textbf{Method} & \makecell{\textbf{Subject}\\\textbf{Consistency\textuparrow}} & \makecell{\textbf{Background}\\\textbf{Consistency\textuparrow}} & \makecell{\textbf{Aesthetic}\\\textbf{Quality\textuparrow}} & \makecell{\textbf{Video-text}\\\textbf{Alignment\textuparrow}}\\
		\midrule
		MEVG & 0.782 & 0.853 & 0.324 & 0.176 \\
		LTX-Video & \underline{0.833} & \underline{0.894} & \textbf{0.483} & \textbf{0.231} \\
		\rowcolor{gray!15}
		Ours & \textbf{0.944} & \textbf{0.951} & \underline{0.448} & \underline{0.180} \\
		\bottomrule
	\end{tabular}
	\caption{\textbf{Comparison on multi-shot video generation.} 
		The baselines do not support video-conditioned generation, so we are unable to report the reference-based metrics.
	}
	\label{tab:multishot}
\end{table}

\paragraph{Implementation Details.}
We adopt LTX-Video-V0.9.0~\cite{hacohen2024ltx}, a text-to-video DiT, as our base model. Its VAE provides a high compression ratio, which facilitates efficient experimentation. Although newer versions of LTX-Video support video extension, we intentionally choose the pure text-to-video variant to demonstrate the effectiveness of our method in enabling video conditioning on top of a non-autoregressive baseline.
We train the model in three stages. In the first stage, we train the lightweight FlexFormer autoencoder from scratch for 20K steps in a batch size of 128, with MSE loss. In the second stage, we freeze the FlexFormer and train the DiT for 30K steps, also using a batch size of 128. For each mini-batch, we randomly sample one of the tasks: prediction, interpolation or retrodiction. The results in Table~\ref{tab:singleshot} are obtained using a linearly-scaled compression strategy. In the third stage, we train the model on multi-shot data for 10K steps. A temporal gap of 20 latent frames is empirically chosen to enable multi-shot capabilities.

\begin{table}
	\scriptsize
	\centering
	\begin{tabular}{c|cccc}
		\toprule
		\textbf{Method} & \textbf{PSNR\textuparrow} & \textbf{SSIM\textuparrow} & \textbf{LPIPS\textdownarrow} \\
		\midrule
		M-RoPE + Multiple Query Tokens & 13.59 & 0.505 & 0.769 \\
		M-RoPE + Single Query Token & 17.95 & 0.611 & 0.470 \\
		I-RoPE + Single Query Token & \textbf{21.01} & \textbf{0.712} & \textbf{0.244} \\
		I-RoPE + MLP Query & \underline{18.45} & \underline{0.640} & \underline{0.407} \\
		\bottomrule
	\end{tabular}
	\caption{\textbf{Ablation of our FlexFormer autoencoder components on video reconstruction.}
		Videos are reconstructed by cascading our FlexFormer and the VAE from the base model. The performance is affected by the VAE.
	}
	\label{tab:ablation-autoencoder}
\end{table}

\subsection{Main Results}

\paragraph{Qualitative Comparison.}
Figure~\ref{fig:single_shot} presents qualitative comparisons on video prediction, interpolation and retrodiction tasks. In video prediction, block-autoregressive methods suffer from subject identity drift, background inconsistency and color flickering. Without specific mechanisms for long-range consistency, CausVid tends to produce frames that become increasingly bright and oversaturated, while SkyReel-V2 generates frames that gradually darken. The 13B model FramePack also exhibits noticeable color flickering across video segments. LTX-Video occasionally suffers from flickering artifacts. In contrast, our method generates the most temporally stable and visually consistent videos.
For interpolation, FramePack sometimes fails to produce smooth transitions, while LTX-Video outputs appear blurry in some cases. Our method generates intermediate frames with clearer structure and smoother motion.
For retrodiction, our method also yields the most natural and coherent results across frames, effectively recovering plausible content for preceding segments.

\begin{table*}[tb]
	\scriptsize
	\centering
	\begin{tabular}{c|c|ccccccc}
		\toprule
		\makebox[0.02\textwidth]{\makecell{\textbf{Compres.}\\\textbf{Strategy}}} & \makebox[0.018\textwidth]{\makecell{\textbf{Compres.}\\\textbf{Ratio}}} & \textbf{PSNR\textuparrow} & \textbf{LPIPS\textdownarrow} & \makecell{\textbf{Subject}\\\textbf{Consistency\textuparrow}} & \makecell{\textbf{Background}\\\textbf{Consistency\textuparrow}} & \makecell{\textbf{Motion}\\\textbf{Smoothness\textuparrow}} & \makecell{\textbf{Aesthetic}\\\textbf{Quality\textuparrow}} & \makecell{\textbf{Video-text}\\\textbf{Alignment\textuparrow}}\\
		\midrule
		\multirow{2}{*}{Uniform} & 8 & \makebox[0.098\textwidth]{12.88/14.44/\textbf{13.79}} & \makebox[0.097\textwidth]{0.480/0.399/0.427} & \makebox[0.097\textwidth]{\underline{0.951}/\underline{0.944}/0.950} & \makebox[0.097\textwidth]{\underline{0.959}/\textbf{0.955}/0.958} & \makebox[0.097\textwidth]{0.992/0.992/0.992} & \makebox[0.097\textwidth]{0.421/0.422/0.421} & \makebox[0.097\textwidth]{\textbf{0.241}/0.223/\textbf{0.229}} \\
		~ & 4 & \makebox[0.097\textwidth]{12.21/13.90/12.34} & \makebox[0.097\textwidth]{0.526/0.436/0.512} & \makebox[0.097\textwidth]{0.925/0.926/0.931} & \makebox[0.097\textwidth]{0.937/0.940/0.939} & \makebox[0.097\textwidth]{0.992/0.993/0.992} & \makebox[0.097\textwidth]{0.404/0.408/0.404} & \makebox[0.097\textwidth]{0.232/0.221/0.224} \\
		\hline
		\multirow{2}{*}{Linear} & \makebox[0.05\textwidth]{5.4: 16 $\rightarrow$ 1} & \makebox[0.097\textwidth]{\textbf{14.12}/\underline{15.06}/\underline{13.78}} & \makebox[0.097\textwidth]{\textbf{0.400}/\underline{0.354}/\underline{0.418}} & \makebox[0.097\textwidth]{0.919/0.919/0.929} & \makebox[0.097\textwidth]{0.940/0.942/0.946} & \makebox[0.097\textwidth]{0.991/0.991/0.991} & \makebox[0.097\textwidth]{0.414/0.420/0.412} & \makebox[0.097\textwidth]{0.239/\underline{0.228}/\textbf{0.229}} \\
		~ & \makebox[0.05\textwidth]{3.4: 8 $\rightarrow$ 1} & \makebox[0.097\textwidth]{\underline{13.39}/\textbf{15.52}/13.76} & \makebox[0.097\textwidth]{\underline{0.427}/\textbf{0.321}/\textbf{0.406}} & \makebox[0.097\textwidth]{0.921/\textbf{0.926}/\underline{0.955}} & \makebox[0.097\textwidth]{0.931/0.939/\underline{0.958}} & \makebox[0.097\textwidth]{\underline{0.994}/\textbf{0.994}/\textbf{0.995}} & \makebox[0.097\textwidth]{0.426/0.434/0.428} & \makebox[0.097\textwidth]{\textbf{0.241}/\textbf{0.229}/0.228}\\
		\hline
		\multirow{2}{*}{Log} & \makebox[0.05\textwidth]{9.7: 16 $\rightarrow$ 1} & \makebox[0.097\textwidth]{11.74/14.82/12.11} & \makebox[0.097\textwidth]{0.530/0.359/0.496} & \makebox[0.097\textwidth]{0.926/0.921/0.920} & \makebox[0.097\textwidth]{0.942/0.940/0.941} & \makebox[0.097\textwidth]{0.990/0.991/0.991} & \makebox[0.097\textwidth]{\underline{0.452}/\underline{0.458}/\underline{0.450}} & \makebox[0.097\textwidth]{0.228/\underline{0.228}/0.226} \\
		~ & \makebox[0.05\textwidth]{5.8: 8 $\rightarrow$ 1} & \makebox[0.098\textwidth]{10.91/14.22/12.24} & \makebox[0.097\textwidth]{0.567/0.379/0.482} & \makebox[0.097\textwidth]{\textbf{0.970}/0.935/\textbf{0.963}} & \makebox[0.097\textwidth]{\textbf{0.963}/\underline{0.946}/\textbf{0.960}} & \makebox[0.097\textwidth]{\textbf{0.995}/\textbf{0.994}/\underline{0.994}} & \makebox[0.097\textwidth]{\textbf{0.461}/\textbf{0.464}/\textbf{0.466}} & \makebox[0.097\textwidth]{0.223/0.227/0.223} \\
		\bottomrule
	\end{tabular}
	\caption{\textbf{Ablation of compression strategy on DiT performance.}
		The first two rows show that a smaller compression ratio does not always bring about better generation performance. For linear and log compression strategies, we first give the overall compression ratio then the compression ratios at double ends.
	}
	\label{tab:ablation-dit}
\end{table*}

\paragraph{Quantitative Comparison.}
Table~\ref{tab:singleshot} presents a quantitative comparison against state-of-the-art DiT-based models on video prediction, interpolation and retrodiction tasks. CausVid and SkyReel-V2 are block-autoregressive methods and therefore do not support interpolation or retrodiction. FramePack does not support retrodiction either. In contrast, our method supports all three tasks as well as multi-shot video generation, owing to the flexibility of our model architecture and training paradigm. Our model outperforms all baselines in terms of continuity between the conditioning and generated video, as measured by PSNR, LPIPS, Subject Consistency and Background Consistency. For non-reference metrics---Aesthetic Quality and Video-text Alignment---our method achieves performance comparable to the 13B FramePack model, despite using only around 2B parameters and being trained exclusively on open-source data. We observe a general trend that longer context length leads to improved temporal consistency and smoother transitions from the conditioning video. Among the five models listed in Table~\ref{tab:singleshot}, increasing context length appears correlated with better metric scores. Notably, context compression does not necessarily compromise performance: although LTX-Video-V0.9.6 consumes the full conditioning video without compression, it still underperforms our model. This highlights the effectiveness of our method.

Table~\ref{tab:multishot} compares our method with multi-shot generation baselines. Generating individual shots separately often results in inconsistency between the shots.
MEVG addresses this by using DDIM inversion of the previous shot to initialize the noise for the next one. However, this approach may struggle in scenarios requiring structural changes between shots. Our method achieves the highest Subject Consistency and Background Consistency among all methods. However, it yields slightly lower Aesthetic Quality score and Video-text Alignment scores compared to the base model LTX-Video. We attribute this to two possible factors: the relatively lower quality of our training data and the fact that our model tends to prioritize coherence with preceding video segments over strict adherence to the input prompt.

\begin{figure}
	\centering
	\includegraphics[width=\linewidth]{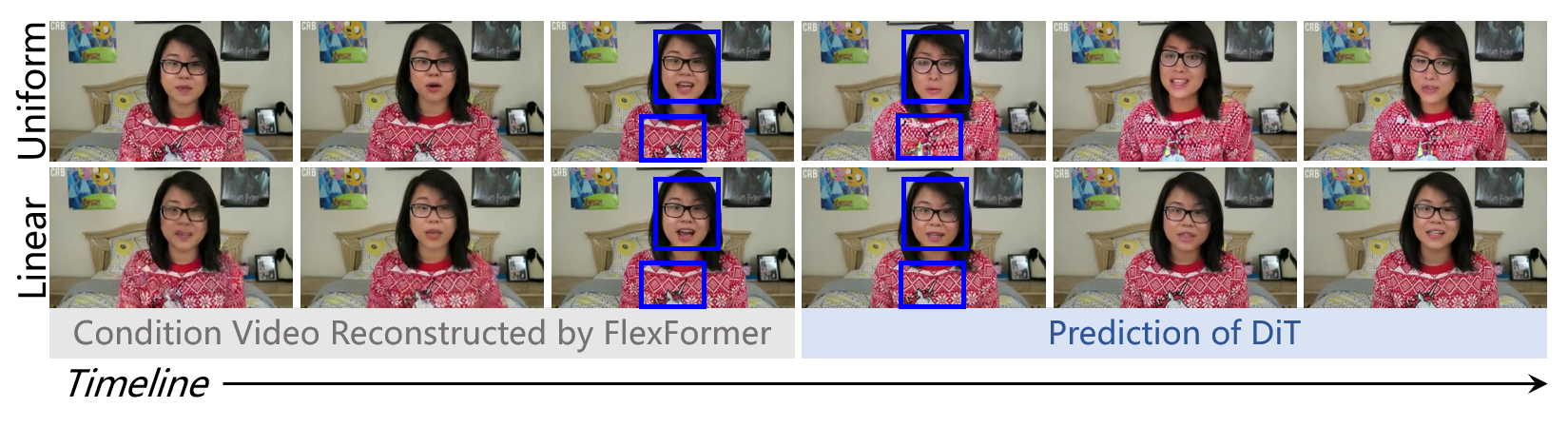}
	\caption{\textbf{Comparison of different compression strategies.} Though uniform strategy achieves a better reconstruction performance, the linear strategy makes it easier for DiT to get a smooth transition, shown by the better consistency of ID and details, as marked by \textcolor{blue}{blue} boxes.
	}
	\label{fig:compression_dit}
\end{figure}

\begin{figure}
	\centering
	\includegraphics[width=\linewidth]{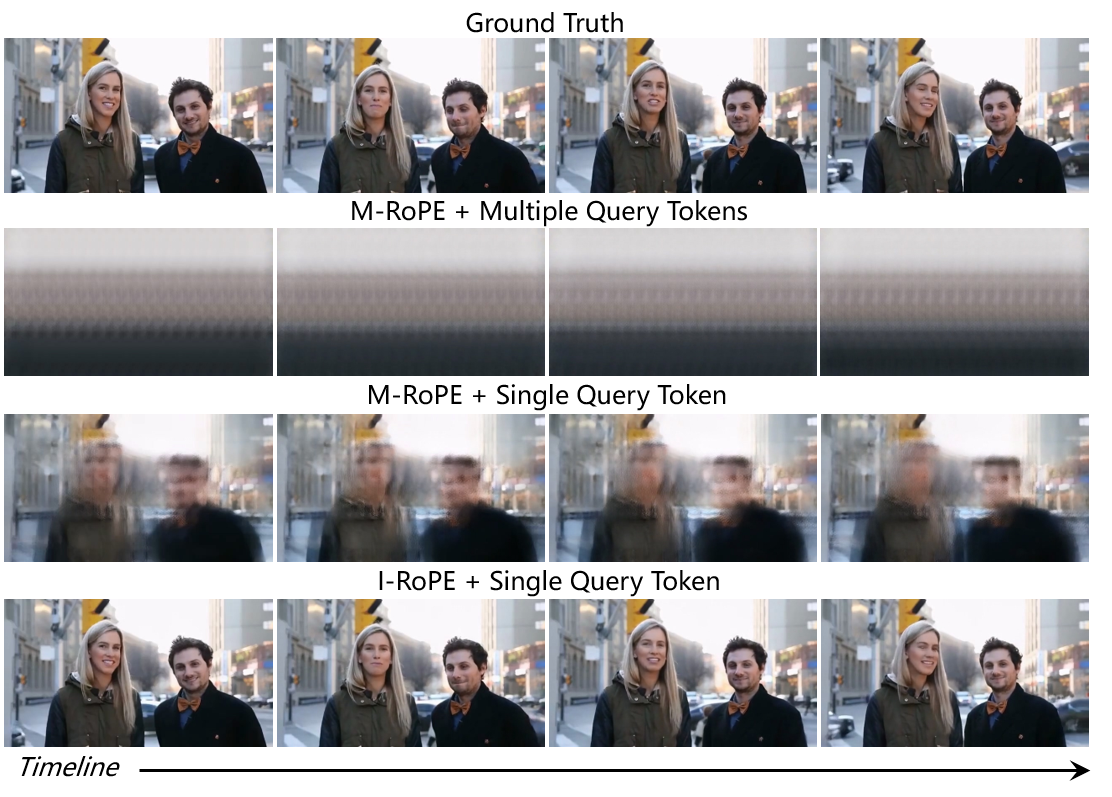}
	\caption{\textbf{Ablation of FlexFormer components.} FlexFormer is cascaded with VAE to reconstruct the videos.
	}
	\label{fig:autoencoder}
\end{figure}

\subsection{Ablation Study}

\paragraph{Ablation of FlexFormer Components.}
Table~\ref{tab:ablation-autoencoder} presents the results of ablating components of FlexFormer. The first setting resembles the vanilla Q-Former except that it replaces cross-attention with self-attention and incorporates M-RoPE to differentiate the positions of video features and learnable query tokens, where the queries are treated as text tokens. When replacing the multiple query tokens with single learnable token, we can assign adaptive query length to videos of different lengths. I-RoPE introduces the spatial correspondence between video tokens and compressed tokens. After adding an MLP in front of the query tokens---taking RoPE embedding of as input---leads to a slight performance drop. We hypothesize that I-RoPE alone is sufficient for encoding spatial information, and the additional MLP may introduce unnecessary complexity or redundancy.

\paragraph{Effect of Compression Strategy on DiT Performance.}
Table~\ref{tab:ablation-dit} shows that the performance of DiT does not necessarily correlate with the reconstruction quality of the FlexFormer autoencoder. When applying linear or log compression strategy, the frames adjacent to the current frames are assigned a lower compression ratio. This design benefits video continuity, as reflected by improvements in PSNR, LPIPS and visual results shown in Figure~\ref{fig:compression_dit}. Among the tested strategies, linear compression offers the most balanced trade-off between performance and efficiency. Hence, we adopt it for the results reported in Table~\ref{tab:singleshot}. Interestingly, using a lower overall compression ratio does not always bring about better DiT performance, suggesting the presence of redundancy in the conditioning feature.

\section{Conclusion}
In this paper, we proposed an efficient and flexible architecture for long video generation, capable of supporting video continuation in any temporal direction as well as multi-shot video generation. Compared to vanilla DiT, our method exhibits greater scalability and has the potential to generate significantly longer videos. In future work, we aim to explore architecture that can adaptively extract and integrate relevant information from previously generated video segments, enabling more coherent and context-aware generation over extended time spans.

\bibliography{aaai2026}

\begin{thebibliography}{96}
\providecommand{\natexlab}[1]{#1}

\bibitem[{Albergo and Vanden-Eijnden(2022)}]{albergo2022building}
Albergo, M.~S.; and Vanden-Eijnden, E. 2022.
\newblock Building normalizing flows with stochastic interpolants.
\newblock \emph{arXiv preprint arXiv:2209.15571}.

\bibitem[{Bai et~al.(2025)Bai, Chen, Liu, Wang, Ge, Song, Dang, Wang, Wang,
  Tang et~al.}]{bai2025qwen2}
Bai, S.; Chen, K.; Liu, X.; Wang, J.; Ge, W.; Song, S.; Dang, K.; Wang, P.;
  Wang, S.; Tang, J.; et~al. 2025.
\newblock Qwen2. 5-vl technical report.
\newblock \emph{arXiv preprint arXiv:2502.13923}.

\bibitem[{Bansal et~al.(2024)Bansal, Bitton, Yarom, Szpektor, Grover, and
  Chang}]{bansal2024talc}
Bansal, H.; Bitton, Y.; Yarom, M.; Szpektor, I.; Grover, A.; and Chang, K.-W.
  2024.
\newblock Talc: Time-aligned captions for multi-scene text-to-video generation.
\newblock \emph{arXiv preprint arXiv:2405.04682}.

\bibitem[{Bao et~al.(2023)Bao, Nie, Xue, Cao, Li, Su, and Zhu}]{bao2023all}
Bao, F.; Nie, S.; Xue, K.; Cao, Y.; Li, C.; Su, H.; and Zhu, J. 2023.
\newblock All are worth words: A vit backbone for diffusion models.
\newblock In \emph{Proceedings of the IEEE/CVF conference on computer vision
  and pattern recognition}, 22669--22679.

\bibitem[{Bar-Tal et~al.(2024)Bar-Tal, Chefer, Tov, Herrmann, Paiss, Zada,
  Ephrat, Hur, Liu, Raj et~al.}]{bar2024lumiere}
Bar-Tal, O.; Chefer, H.; Tov, O.; Herrmann, C.; Paiss, R.; Zada, S.; Ephrat,
  A.; Hur, J.; Liu, G.; Raj, A.; et~al. 2024.
\newblock Lumiere: A space-time diffusion model for video generation.
\newblock In \emph{SIGGRAPH Asia 2024 Conference Papers}, 1--11.

\bibitem[{Blattmann et~al.(2023{\natexlab{a}})Blattmann, Dockhorn, Kulal,
  Mendelevitch, Kilian, Lorenz, Levi, English, Voleti, Letts
  et~al.}]{blattmann2023stable}
Blattmann, A.; Dockhorn, T.; Kulal, S.; Mendelevitch, D.; Kilian, M.; Lorenz,
  D.; Levi, Y.; English, Z.; Voleti, V.; Letts, A.; et~al. 2023{\natexlab{a}}.
\newblock Stable video diffusion: Scaling latent video diffusion models to
  large datasets.
\newblock \emph{arXiv preprint arXiv:2311.15127}.

\bibitem[{Blattmann et~al.(2023{\natexlab{b}})Blattmann, Rombach, Ling,
  Dockhorn, Kim, Fidler, and Kreis}]{blattmann2023align}
Blattmann, A.; Rombach, R.; Ling, H.; Dockhorn, T.; Kim, S.~W.; Fidler, S.; and
  Kreis, K. 2023{\natexlab{b}}.
\newblock Align your latents: High-resolution video synthesis with latent
  diffusion models.
\newblock In \emph{Proceedings of the IEEE/CVF conference on computer vision
  and pattern recognition}, 22563--22575.

\bibitem[{Chen et~al.(2024{\natexlab{a}})Chen, Mart{\'\i}~Mons{\'o}, Du,
  Simchowitz, Tedrake, and Sitzmann}]{chen2024diffusion}
Chen, B.; Mart{\'\i}~Mons{\'o}, D.; Du, Y.; Simchowitz, M.; Tedrake, R.; and
  Sitzmann, V. 2024{\natexlab{a}}.
\newblock Diffusion forcing: Next-token prediction meets full-sequence
  diffusion.
\newblock \emph{Advances in Neural Information Processing Systems}, 37:
  24081--24125.

\bibitem[{Chen et~al.(2025{\natexlab{a}})Chen, Lin, Yang, Lin, Zhu, Fan, Zhang,
  Chen, Chen, Ma et~al.}]{chen2025skyreels}
Chen, G.; Lin, D.; Yang, J.; Lin, C.; Zhu, J.; Fan, M.; Zhang, H.; Chen, S.;
  Chen, Z.; Ma, C.; et~al. 2025{\natexlab{a}}.
\newblock Skyreels-v2: Infinite-length film generative model.
\newblock \emph{arXiv preprint arXiv:2504.13074}.

\bibitem[{Chen et~al.(2023{\natexlab{a}})Chen, Xia, He, Zhang, Cun, Yang, Xing,
  Liu, Chen, Wang et~al.}]{chen2023videocrafter1}
Chen, H.; Xia, M.; He, Y.; Zhang, Y.; Cun, X.; Yang, S.; Xing, J.; Liu, Y.;
  Chen, Q.; Wang, X.; et~al. 2023{\natexlab{a}}.
\newblock Videocrafter1: Open diffusion models for high-quality video
  generation.
\newblock \emph{arXiv preprint arXiv:2310.19512}.

\bibitem[{Chen et~al.(2024{\natexlab{b}})Chen, Zhang, Cun, Xia, Wang, Weng, and
  Shan}]{chen2024videocrafter2}
Chen, H.; Zhang, Y.; Cun, X.; Xia, M.; Wang, X.; Weng, C.; and Shan, Y.
  2024{\natexlab{b}}.
\newblock Videocrafter2: Overcoming data limitations for high-quality video
  diffusion models.
\newblock In \emph{Proceedings of the IEEE/CVF Conference on Computer Vision
  and Pattern Recognition}, 7310--7320.

\bibitem[{Chen et~al.(2025{\natexlab{b}})Chen, Long, An, Qiu, Yao, Luo, and
  Mei}]{chen2025ouroboros}
Chen, J.; Long, F.; An, J.; Qiu, Z.; Yao, T.; Luo, J.; and Mei, T.
  2025{\natexlab{b}}.
\newblock Ouroboros-Diffusion: Exploring Consistent Content Generation in
  Tuning-free Long Video Diffusion.
\newblock \emph{arXiv preprint arXiv:2501.09019}.

\bibitem[{Chen et~al.(2024{\natexlab{c}})Chen, Siarohin, Menapace, Deyneka,
  Chao, Jeon, Fang, Lee, Ren, Yang et~al.}]{chen2024panda}
Chen, T.-S.; Siarohin, A.; Menapace, W.; Deyneka, E.; Chao, H.-w.; Jeon, B.~E.;
  Fang, Y.; Lee, H.-Y.; Ren, J.; Yang, M.-H.; et~al. 2024{\natexlab{c}}.
\newblock Panda-70m: Captioning 70m videos with multiple cross-modality
  teachers.
\newblock In \emph{Proceedings of the IEEE/CVF Conference on Computer Vision
  and Pattern Recognition}, 13320--13331.

\bibitem[{Chen et~al.(2023{\natexlab{b}})Chen, Wang, Zhang, Zhuang, Ma, Yu,
  Wang, Lin, Qiao, and Liu}]{chen2023seine}
Chen, X.; Wang, Y.; Zhang, L.; Zhuang, S.; Ma, X.; Yu, J.; Wang, Y.; Lin, D.;
  Qiao, Y.; and Liu, Z. 2023{\natexlab{b}}.
\newblock Seine: Short-to-long video diffusion model for generative transition
  and prediction.
\newblock In \emph{The Twelfth International Conference on Learning
  Representations}.

\bibitem[{Dalal et~al.(2025)Dalal, Koceja, Hussein, Xu, Zhao, Song, Han,
  Cheung, Kautz, Guestrin et~al.}]{dalal2025one}
Dalal, K.; Koceja, D.; Hussein, G.; Xu, J.; Zhao, Y.; Song, Y.; Han, S.;
  Cheung, K.~C.; Kautz, J.; Guestrin, C.; et~al. 2025.
\newblock One-minute video generation with test-time training.
\newblock \emph{arXiv preprint arXiv:2504.05298}.

\bibitem[{Esser et~al.(2024)Esser, Kulal, Blattmann, Entezari, M{\"u}ller,
  Saini, Levi, Lorenz, Sauer, Boesel et~al.}]{esser2024scaling}
Esser, P.; Kulal, S.; Blattmann, A.; Entezari, R.; M{\"u}ller, J.; Saini, H.;
  Levi, Y.; Lorenz, D.; Sauer, A.; Boesel, F.; et~al. 2024.
\newblock Scaling rectified flow transformers for high-resolution image
  synthesis.
\newblock In \emph{Forty-first international conference on machine learning}.

\bibitem[{Fang et~al.(2025)Fang, Ma, Chen, Zhou, and Qi}]{fang2025inflvg}
Fang, X.; Ma, L.; Chen, Z.; Zhou, M.; and Qi, G.-j. 2025.
\newblock InfLVG: Reinforce Inference-Time Consistent Long Video Generation
  with GRPO.
\newblock \emph{arXiv preprint arXiv:2505.17574}.

\bibitem[{Gao et~al.(2024{\natexlab{a}})Gao, Shi, Zhang, Wang, and
  Xiao}]{gao2024vid}
Gao, K.; Shi, J.; Zhang, H.; Wang, C.; and Xiao, J. 2024{\natexlab{a}}.
\newblock Vid-gpt: Introducing gpt-style autoregressive generation in video
  diffusion models.
\newblock \emph{arXiv preprint arXiv:2406.10981}.

\bibitem[{Gao et~al.(2024{\natexlab{b}})Gao, Shi, Zhang, Wang, Xiao, and
  Chen}]{gao2024ca2}
Gao, K.; Shi, J.; Zhang, H.; Wang, C.; Xiao, J.; and Chen, L.
  2024{\natexlab{b}}.
\newblock Ca2-VDM: Efficient Autoregressive Video Diffusion Model with Causal
  Generation and Cache Sharing.
\newblock \emph{arXiv preprint arXiv:2411.16375}.

\bibitem[{Gu et~al.(2023)Gu, Wang, Zhao, Lu, Zhang, Wu, Xu, Zhang, Jiang, and
  Xu}]{gu2023reuse}
Gu, J.; Wang, S.; Zhao, H.; Lu, T.; Zhang, X.; Wu, Z.; Xu, S.; Zhang, W.;
  Jiang, Y.-G.; and Xu, H. 2023.
\newblock Reuse and diffuse: Iterative denoising for text-to-video generation.
\newblock \emph{arXiv preprint arXiv:2309.03549}.

\bibitem[{Gu, Mao, and Shou(2025)}]{gu2025long}
Gu, Y.; Mao, W.; and Shou, M.~Z. 2025.
\newblock Long-context autoregressive video modeling with next-frame
  prediction.
\newblock \emph{arXiv preprint arXiv:2503.19325}.

\bibitem[{Guo et~al.(2023)Guo, Yang, Rao, Liang, Wang, Qiao, Agrawala, Lin, and
  Dai}]{guo2023animatediff}
Guo, Y.; Yang, C.; Rao, A.; Liang, Z.; Wang, Y.; Qiao, Y.; Agrawala, M.; Lin,
  D.; and Dai, B. 2023.
\newblock Animatediff: Animate your personalized text-to-image diffusion models
  without specific tuning.
\newblock \emph{arXiv preprint arXiv:2307.04725}.

\bibitem[{Guo et~al.(2025)Guo, Yang, Yang, Ma, Lin, Yang, Lin, and
  Jiang}]{guo2025long}
Guo, Y.; Yang, C.; Yang, Z.; Ma, Z.; Lin, Z.; Yang, Z.; Lin, D.; and Jiang, L.
  2025.
\newblock Long context tuning for video generation.
\newblock \emph{arXiv preprint arXiv:2503.10589}.

\bibitem[{HaCohen et~al.(2024)HaCohen, Chiprut, Brazowski, Shalem, Moshe,
  Richardson, Levin, Shiran, Zabari, Gordon et~al.}]{hacohen2024ltx}
HaCohen, Y.; Chiprut, N.; Brazowski, B.; Shalem, D.; Moshe, D.; Richardson, E.;
  Levin, E.; Shiran, G.; Zabari, N.; Gordon, O.; et~al. 2024.
\newblock Ltx-video: Realtime video latent diffusion.
\newblock \emph{arXiv preprint arXiv:2501.00103}.

\bibitem[{Harvey et~al.(2022)Harvey, Naderiparizi, Masrani, Weilbach, and
  Wood}]{harvey2022flexible}
Harvey, W.; Naderiparizi, S.; Masrani, V.; Weilbach, C.; and Wood, F. 2022.
\newblock Flexible diffusion modeling of long videos.
\newblock \emph{Advances in Neural Information Processing Systems}, 35:
  27953--27965.

\bibitem[{Henschel et~al.(2024)Henschel, Khachatryan, Hayrapetyan, Poghosyan,
  Tadevosyan, Wang, Navasardyan, and Shi}]{henschel2024streamingt2v}
Henschel, R.; Khachatryan, L.; Hayrapetyan, D.; Poghosyan, H.; Tadevosyan, V.;
  Wang, Z.; Navasardyan, S.; and Shi, H. 2024.
\newblock Streamingt2v: Consistent, dynamic, and extendable long video
  generation from text.
\newblock \emph{arXiv preprint arXiv:2403.14773}.

\bibitem[{Ho et~al.(2022{\natexlab{a}})Ho, Chan, Saharia, Whang, Gao,
  Gritsenko, Kingma, Poole, Norouzi, Fleet, and Salimans}]{ho2022imagen}
Ho, J.; Chan, W.; Saharia, C.; Whang, J.; Gao, R.; Gritsenko, A.; Kingma,
  D.~P.; Poole, B.; Norouzi, M.; Fleet, D.~J.; and Salimans, T.
  2022{\natexlab{a}}.
\newblock Imagen Video: High Definition Video Generation with Diffusion Models.
\newblock \emph{arXiv preprint arXiv: 2210.02303}.

\bibitem[{Ho et~al.(2022{\natexlab{b}})Ho, Salimans, Gritsenko, Chan, Norouzi,
  and Fleet}]{ho2022video}
Ho, J.; Salimans, T.; Gritsenko, A.; Chan, W.; Norouzi, M.; and Fleet, D.~J.
  2022{\natexlab{b}}.
\newblock Video diffusion models.
\newblock \emph{Advances in Neural Information Processing Systems}, 35:
  8633--8646.

\bibitem[{Hu et~al.(2024)Hu, Hu, Song, Huang, Wang, Zhou, Liu, Ma, and
  Sun}]{hu2024acdit}
Hu, J.; Hu, S.; Song, Y.; Huang, Y.; Wang, M.; Zhou, H.; Liu, Z.; Ma, W.-Y.;
  and Sun, M. 2024.
\newblock ACDiT: Interpolating Autoregressive Conditional Modeling and
  Diffusion Transformer.
\newblock \emph{arXiv preprint arXiv:2412.07720}.

\bibitem[{Huang et~al.(2025)Huang, Li, He, Zhou, and Shechtman}]{huang2025self}
Huang, X.; Li, Z.; He, G.; Zhou, M.; and Shechtman, E. 2025.
\newblock Self Forcing: Bridging the Train-Test Gap in Autoregressive Video
  Diffusion.
\newblock \emph{arXiv preprint arXiv:2506.08009}.

\bibitem[{Huang et~al.(2024)Huang, He, Yu, Zhang, Si, Jiang, Zhang, Wu, Jin,
  Chanpaisit et~al.}]{huang2024vbench}
Huang, Z.; He, Y.; Yu, J.; Zhang, F.; Si, C.; Jiang, Y.; Zhang, Y.; Wu, T.;
  Jin, Q.; Chanpaisit, N.; et~al. 2024.
\newblock Vbench: Comprehensive benchmark suite for video generative models.
\newblock In \emph{Proceedings of the IEEE/CVF Conference on Computer Vision
  and Pattern Recognition}, 21807--21818.

\bibitem[{Jin et~al.(2024)Jin, Sun, Li, Xu, Jiang, Zhuang, Huang, Song, Mu, and
  Lin}]{jin2024pyramidal}
Jin, Y.; Sun, Z.; Li, N.; Xu, K.; Jiang, H.; Zhuang, N.; Huang, Q.; Song, Y.;
  Mu, Y.; and Lin, Z. 2024.
\newblock Pyramidal flow matching for efficient video generative modeling.
\newblock \emph{arXiv preprint arXiv:2410.05954}.

\bibitem[{Kang, Kothandaraman, and Lin(2025)}]{kang2025text2story}
Kang, T.; Kothandaraman, D.; and Lin, M.~C. 2025.
\newblock Text2story: Advancing video storytelling with text guidance.
\newblock \emph{arXiv preprint arXiv:2503.06310}.

\bibitem[{Kara et~al.(2025)Kara, Singh, Liu, Ceylan, Rehg, and
  Hinz}]{kara2025shotadapter}
Kara, O.; Singh, K.~K.; Liu, F.; Ceylan, D.; Rehg, J.~M.; and Hinz, T. 2025.
\newblock ShotAdapter: Text-to-Multi-Shot Video Generation with Diffusion
  Models.
\newblock \emph{arXiv preprint arXiv:2505.07652}.

\bibitem[{Kim et~al.(2025{\natexlab{a}})Kim, Kang, Choi, and Han}]{kim2025fifo}
Kim, J.; Kang, J.; Choi, J.; and Han, B. 2025{\natexlab{a}}.
\newblock FIFO-Diffusion: Generating Infinite Videos from Text without
  Training.
\newblock \emph{Advances in Neural Information Processing Systems}, 37:
  89834--89868.

\bibitem[{Kim et~al.(2025{\natexlab{b}})Kim, Oh, Wang, Lee, and
  Shin}]{kim2025tuning}
Kim, S.; Oh, S.~W.; Wang, J.-H.; Lee, J.-Y.; and Shin, J. 2025{\natexlab{b}}.
\newblock Tuning-Free Multi-Event Long Video Generation via Synchronized
  Coupled Sampling.
\newblock \emph{arXiv preprint arXiv:2503.08605}.

\bibitem[{Kong et~al.(2024)Kong, Tian, Zhang, Min, Dai, Zhou, Xiong, Li, Wu,
  Zhang et~al.}]{kong2024hunyuanvideo}
Kong, W.; Tian, Q.; Zhang, Z.; Min, R.; Dai, Z.; Zhou, J.; Xiong, J.; Li, X.;
  Wu, B.; Zhang, J.; et~al. 2024.
\newblock Hunyuanvideo: A systematic framework for large video generative
  models.
\newblock \emph{arXiv preprint arXiv:2412.03603}.

\bibitem[{Li et~al.(2023)Li, Li, Savarese, and Hoi}]{li2023blip}
Li, J.; Li, D.; Savarese, S.; and Hoi, S. 2023.
\newblock Blip-2: Bootstrapping language-image pre-training with frozen image
  encoders and large language models.
\newblock In \emph{International conference on machine learning}, 19730--19742.
  PMLR.

\bibitem[{Li et~al.(2024{\natexlab{a}})Li, Beluch, Keuper, Zhang, and
  Khoreva}]{li2024vstar}
Li, Y.; Beluch, W.; Keuper, M.; Zhang, D.; and Khoreva, A. 2024{\natexlab{a}}.
\newblock Vstar: Generative temporal nursing for longer dynamic video
  synthesis.
\newblock \emph{arXiv preprint arXiv:2403.13501}.

\bibitem[{Li et~al.(2024{\natexlab{b}})Li, Hu, Liu, Zhou, Choi, Meng, Guo, Li,
  Ling, and Wei}]{li2024arlon}
Li, Z.; Hu, S.; Liu, S.; Zhou, L.; Choi, J.; Meng, L.; Guo, X.; Li, J.; Ling,
  H.; and Wei, F. 2024{\natexlab{b}}.
\newblock Arlon: Boosting diffusion transformers with autoregressive models for
  long video generation.
\newblock \emph{arXiv preprint arXiv:2410.20502}.

\bibitem[{Lin et~al.(2024)Lin, Ge, Cheng, Li, Zhu, Wang, He, Ye, Yuan, Chen
  et~al.}]{lin2024open}
Lin, B.; Ge, Y.; Cheng, X.; Li, Z.; Zhu, B.; Wang, S.; He, X.; Ye, Y.; Yuan,
  S.; Chen, L.; et~al. 2024.
\newblock Open-sora plan: Open-source large video generation model.
\newblock \emph{arXiv preprint arXiv:2412.00131}.

\bibitem[{Lin et~al.(2025)Lin, Yang, He, Jiang, Ren, Xia, Zhao, Xiao, and
  Jiang}]{lin2025autoregressive}
Lin, S.; Yang, C.; He, H.; Jiang, J.; Ren, Y.; Xia, X.; Zhao, Y.; Xiao, X.; and
  Jiang, L. 2025.
\newblock Autoregressive Adversarial Post-Training for Real-Time Interactive
  Video Generation.
\newblock \emph{arXiv preprint arXiv:2506.09350}.

\bibitem[{Lipman et~al.(2022)Lipman, Chen, Ben-Hamu, Nickel, and
  Le}]{lipman2022flow}
Lipman, Y.; Chen, R.~T.; Ben-Hamu, H.; Nickel, M.; and Le, M. 2022.
\newblock Flow matching for generative modeling.
\newblock \emph{arXiv preprint arXiv:2210.02747}.

\bibitem[{Liu et~al.(2025)Liu, Li, Liu, Li, Wang, Li, Qin, Liu, Xin, Li
  et~al.}]{liu2025lumina}
Liu, D.; Li, S.; Liu, Y.; Li, Z.; Wang, K.; Li, X.; Qin, Q.; Liu, Y.; Xin, Y.;
  Li, Z.; et~al. 2025.
\newblock Lumina-Video: Efficient and Flexible Video Generation with
  Multi-scale Next-DiT.
\newblock \emph{arXiv preprint arXiv:2502.06782}.

\bibitem[{Liu, Gong, and Liu(2022)}]{liu2022flow}
Liu, X.; Gong, C.; and Liu, Q. 2022.
\newblock Flow straight and fast: Learning to generate and transfer data with
  rectified flow.
\newblock \emph{arXiv preprint arXiv:2209.03003}.

\bibitem[{Long et~al.(2024)Long, Qiu, Yao, and Mei}]{long2024videostudio}
Long, F.; Qiu, Z.; Yao, T.; and Mei, T. 2024.
\newblock VideoStudio: Generating Consistent-Content and Multi-Scene Videos.
\newblock In \emph{European Conference on Computer Vision}, 468--485. Springer.

\bibitem[{Lu et~al.()Lu, Liang, Zhu, and Yang}]{lufreelong}
Lu, Y.; Liang, Y.; Zhu, L.; and Yang, Y. ????
\newblock FreeLong: Training-Free Long Video Generation with SpectralBlend
  Temporal Attention.
\newblock In \emph{The Thirty-eighth Annual Conference on Neural Information
  Processing Systems}.

\bibitem[{Ma et~al.(2025)Ma, Huang, Yan, Chen, Duan, Yin, Wan, Ming, Song, Chen
  et~al.}]{ma2025step}
Ma, G.; Huang, H.; Yan, K.; Chen, L.; Duan, N.; Yin, S.; Wan, C.; Ming, R.;
  Song, X.; Chen, X.; et~al. 2025.
\newblock Step-video-t2v technical report: The practice, challenges, and future
  of video foundation model.
\newblock \emph{arXiv preprint arXiv:2502.10248}.

\bibitem[{Oh et~al.(2024)Oh, Jeong, Kim, Byeon, Kim, Kim, and Kim}]{oh2024mevg}
Oh, G.; Jeong, J.; Kim, S.; Byeon, W.; Kim, J.; Kim, S.; and Kim, S. 2024.
\newblock Mevg: Multi-event video generation with text-to-video models.
\newblock In \emph{European Conference on Computer Vision}, 401--418. Springer.

\bibitem[{Ouyang et~al.(2024)Ouyang, Zhao, Wang et~al.}]{ouyang2024flexifilm}
Ouyang, Y.; Zhao, H.; Wang, G.; et~al. 2024.
\newblock Flexifilm: Long video generation with flexible conditions.
\newblock \emph{arXiv preprint arXiv:2404.18620}.

\bibitem[{Peebles and Xie(2023)}]{peebles2023scalable}
Peebles, W.; and Xie, S. 2023.
\newblock Scalable diffusion models with transformers.
\newblock In \emph{Proceedings of the IEEE/CVF international conference on
  computer vision}, 4195--4205.

\bibitem[{Polyak et~al.(2024)Polyak, Zohar, Brown, Tjandra, Sinha, Lee, Vyas,
  Shi, Ma, Chuang et~al.}]{polyak2024movie}
Polyak, A.; Zohar, A.; Brown, A.; Tjandra, A.; Sinha, A.; Lee, A.; Vyas, A.;
  Shi, B.; Ma, C.-Y.; Chuang, C.-Y.; et~al. 2024.
\newblock Movie Gen: A Cast of Media Foundation Models.
\newblock \emph{arXiv preprint arXiv:2410.13720}.

\bibitem[{Qi et~al.(2025)Qi, Yuan, Feng, Fang, Liu, Zhou, He, Xie, and
  Zhang}]{qi2025mask}
Qi, T.; Yuan, J.; Feng, W.; Fang, S.; Liu, J.; Zhou, S.; He, Q.; Xie, H.; and
  Zhang, Y. 2025.
\newblock Mask2DiT: Dual Mask-based Diffusion Transformer for Multi-Scene Long
  Video Generation.
\newblock \emph{arXiv preprint arXiv:2503.19881}.

\bibitem[{Qiu et~al.(2023)Qiu, Xia, Zhang, He, Wang, Shan, and
  Liu}]{qiu2023freenoise}
Qiu, H.; Xia, M.; Zhang, Y.; He, Y.; Wang, X.; Shan, Y.; and Liu, Z. 2023.
\newblock Freenoise: Tuning-free longer video diffusion via noise rescheduling.
\newblock \emph{arXiv preprint arXiv:2310.15169}.

\bibitem[{Rombach et~al.(2022)Rombach, Blattmann, Lorenz, Esser, and
  Ommer}]{rombach2022high}
Rombach, R.; Blattmann, A.; Lorenz, D.; Esser, P.; and Ommer, B. 2022.
\newblock High-resolution image synthesis with latent diffusion models.
\newblock In \emph{Proceedings of the IEEE/CVF conference on computer vision
  and pattern recognition}, 10684--10695.

\bibitem[{Ronneberger, Fischer, and Brox(2015)}]{ronneberger2015u}
Ronneberger, O.; Fischer, P.; and Brox, T. 2015.
\newblock U-net: Convolutional networks for biomedical image segmentation.
\newblock In \emph{Medical image computing and computer-assisted
  intervention--MICCAI 2015: 18th international conference, Munich, Germany,
  October 5-9, 2015, proceedings, part III 18}, 234--241. Springer.

\bibitem[{Ruhe et~al.()Ruhe, Heek, Salimans, and Hoogeboom}]{ruherolling}
Ruhe, D.; Heek, J.; Salimans, T.; and Hoogeboom, E. ????
\newblock Rolling Diffusion Models.
\newblock In \emph{Forty-first International Conference on Machine Learning}.

\bibitem[{Savov et~al.(2025)Savov, Kazemi, Zhang, Paudel, Wang, and
  Gool}]{savov2025statespacediffuser0}
Savov, N.; Kazemi, N.; Zhang, D.; Paudel, D.~P.; Wang, X.; and Gool, L.~V.
  2025.
\newblock StateSpaceDiffuser: Bringing Long Context to Diffusion World Models.
\newblock \emph{arXiv preprint arXiv: 2505.22246}.

\bibitem[{Song, Meng, and Ermon(2020)}]{song2020denoising}
Song, J.; Meng, C.; and Ermon, S. 2020.
\newblock Denoising diffusion implicit models.
\newblock \emph{arXiv preprint arXiv:2010.02502}.

\bibitem[{Song et~al.(2025)Song, Chen, Simchowitz, Du, Tedrake, and
  Sitzmann}]{song2025history}
Song, K.; Chen, B.; Simchowitz, M.; Du, Y.; Tedrake, R.; and Sitzmann, V. 2025.
\newblock History-Guided Video Diffusion.
\newblock \emph{arXiv preprint arXiv:2502.06764}.

\bibitem[{Su et~al.(2024)Su, Ahmed, Lu, Pan, Bo, and Liu}]{su2024roformer}
Su, J.; Ahmed, M.; Lu, Y.; Pan, S.; Bo, W.; and Liu, Y. 2024.
\newblock Roformer: Enhanced transformer with rotary position embedding.
\newblock \emph{Neurocomputing}, 568: 127063.

\bibitem[{Sun et~al.(2024)Sun, Zhou, Yuan, Sun, Li, Jia, Adam, Hariharan, Zhao,
  and Liu}]{sun2024video}
Sun, Y.; Zhou, H.; Yuan, L.; Sun, J.~J.; Li, Y.; Jia, X.; Adam, H.; Hariharan,
  B.; Zhao, L.; and Liu, T. 2024.
\newblock Video Creation by Demonstration.
\newblock \emph{arXiv preprint arXiv:2412.09551}.

\bibitem[{Teng et~al.(2025)Teng, Jia, Sun, Li, Li, Tang, Han, Zhang, Zhang, Luo
  et~al.}]{teng2025magi}
Teng, H.; Jia, H.; Sun, L.; Li, L.; Li, M.; Tang, M.; Han, S.; Zhang, T.;
  Zhang, W.; Luo, W.; et~al. 2025.
\newblock MAGI-1: Autoregressive Video Generation at Scale.
\newblock \emph{arXiv preprint arXiv:2505.13211}.

\bibitem[{Voleti, Jolicoeur-Martineau, and Pal(2022)}]{voleti2022mcvd}
Voleti, V.; Jolicoeur-Martineau, A.; and Pal, C. 2022.
\newblock Mcvd-masked conditional video diffusion for prediction, generation,
  and interpolation.
\newblock \emph{Advances in neural information processing systems}, 35:
  23371--23385.

\bibitem[{Wang et~al.(2025{\natexlab{a}})Wang, Ai, Wen, Mao, Xie, Chen, Yu,
  Zhao, Yang, Zeng et~al.}]{wang2025wan}
Wang, A.; Ai, B.; Wen, B.; Mao, C.; Xie, C.-W.; Chen, D.; Yu, F.; Zhao, H.;
  Yang, J.; Zeng, J.; et~al. 2025{\natexlab{a}}.
\newblock Wan: Open and advanced large-scale video generative models.
\newblock \emph{arXiv preprint arXiv:2503.20314}.

\bibitem[{Wang et~al.(2023{\natexlab{a}})Wang, Chen, Song, Ye, Liu, and
  Li}]{wang2023gen}
Wang, F.-Y.; Chen, W.; Song, G.; Ye, H.-J.; Liu, Y.; and Li, H.
  2023{\natexlab{a}}.
\newblock Gen-l-video: Multi-text to long video generation via temporal
  co-denoising.
\newblock \emph{arXiv preprint arXiv:2305.18264}.

\bibitem[{Wang et~al.(2024)Wang, Huang, Ma, Song, Lu, Bian, Li, Liu, and
  Li}]{wang2024zola}
Wang, F.-Y.; Huang, Z.; Ma, Q.; Song, G.; Lu, X.; Bian, W.; Li, Y.; Liu, Y.;
  and Li, H. 2024.
\newblock ZoLA: Zero-Shot Creative Long Animation Generation with Short Video
  Model.
\newblock In \emph{European Conference on Computer Vision}, 329--345. Springer.

\bibitem[{Wang et~al.(2025{\natexlab{b}})Wang, Ma, Liu, Hou, Xu, Wang,
  Juefei-Xu, Luo, Zhang, Hou et~al.}]{wang2025lingen}
Wang, H.; Ma, C.-Y.; Liu, Y.-C.; Hou, J.; Xu, T.; Wang, J.; Juefei-Xu, F.; Luo,
  Y.; Zhang, P.; Hou, T.; et~al. 2025{\natexlab{b}}.
\newblock Lingen: Towards high-resolution minute-length text-to-video
  generation with linear computational complexity.
\newblock In \emph{Proceedings of the Computer Vision and Pattern Recognition
  Conference}, 2578--2588.

\bibitem[{Wang et~al.(2025{\natexlab{c}})Wang, Sheng, Cai, Zhang, Yan, Feng,
  Deng, and Ye}]{wang2025echoshot}
Wang, J.; Sheng, H.; Cai, S.; Zhang, W.; Yan, C.; Feng, Y.; Deng, B.; and Ye,
  J. 2025{\natexlab{c}}.
\newblock EchoShot: Multi-Shot Portrait Video Generation.
\newblock \emph{arXiv preprint arXiv:2506.15838}.

\bibitem[{Wang et~al.(2023{\natexlab{b}})Wang, Yuan, Chen, Zhang, Wang, and
  Zhang}]{wang2023modelscope}
Wang, J.; Yuan, H.; Chen, D.; Zhang, Y.; Wang, X.; and Zhang, S.
  2023{\natexlab{b}}.
\newblock Modelscope text-to-video technical report.
\newblock \emph{arXiv preprint arXiv:2308.06571}.

\bibitem[{Wang et~al.(2025{\natexlab{d}})Wang, Chen, Ma, Zhou, Huang, Wang,
  Yang, He, Yu, Yang et~al.}]{wang2025lavie}
Wang, Y.; Chen, X.; Ma, X.; Zhou, S.; Huang, Z.; Wang, Y.; Yang, C.; He, Y.;
  Yu, J.; Yang, P.; et~al. 2025{\natexlab{d}}.
\newblock Lavie: High-quality video generation with cascaded latent diffusion
  models.
\newblock \emph{International Journal of Computer Vision}, 133(5): 3059--3078.

\bibitem[{Wang et~al.(2023{\natexlab{c}})Wang, He, Li, Li, Yu, Ma, Li, Chen,
  Chen, Wang et~al.}]{wang2023internvid}
Wang, Y.; He, Y.; Li, Y.; Li, K.; Yu, J.; Ma, X.; Li, X.; Chen, G.; Chen, X.;
  Wang, Y.; et~al. 2023{\natexlab{c}}.
\newblock Internvid: A large-scale video-text dataset for multimodal
  understanding and generation.
\newblock \emph{arXiv preprint arXiv:2307.06942}.

\bibitem[{Wu et~al.(2024)Wu, Siarohin, Menapace, Skorokhodov, Fang, Chordia,
  Gilitschenski, and Tulyakov}]{wu2024mind}
Wu, Z.; Siarohin, A.; Menapace, W.; Skorokhodov, I.; Fang, Y.; Chordia, V.;
  Gilitschenski, I.; and Tulyakov, S. 2024.
\newblock Mind the Time: Temporally-Controlled Multi-Event Video Generation.
\newblock \emph{arXiv preprint arXiv:2412.05263}.

\bibitem[{Xiang et~al.(2024)Xiang, Liu, Gu, Gao, Ning, Zha, Feng, Tao, Hao, Shi
  et~al.}]{xiang2024pandora}
Xiang, J.; Liu, G.; Gu, Y.; Gao, Q.; Ning, Y.; Zha, Y.; Feng, Z.; Tao, T.; Hao,
  S.; Shi, Y.; et~al. 2024.
\newblock Pandora: Towards general world model with natural language actions
  and video states.
\newblock \emph{arXiv preprint arXiv:2406.09455}.

\bibitem[{Xiao et~al.(2025{\natexlab{a}})Xiao, Cheng, Qi, Gui, Cen, Ma, Yuille,
  and Jiang}]{xiao2025videoauteur}
Xiao, J.; Cheng, F.; Qi, L.; Gui, L.; Cen, J.; Ma, Z.; Yuille, A.; and Jiang,
  L. 2025{\natexlab{a}}.
\newblock VideoAuteur: Towards Long Narrative Video Generation.
\newblock \emph{arXiv preprint arXiv:2501.06173}.

\bibitem[{Xiao et~al.(2025{\natexlab{b}})Xiao, Lan, Zhou, Ouyang, Yang, Zeng,
  and Pan}]{xiao2025worldmem}
Xiao, Z.; Lan, Y.; Zhou, Y.; Ouyang, W.; Yang, S.; Zeng, Y.; and Pan, X.
  2025{\natexlab{b}}.
\newblock WORLDMEM: Long-term consistent world simulation with memory.
\newblock \emph{arXiv preprint arXiv:2504.12369}.

\bibitem[{Xie et~al.(2024)Xie, Xu, Hong, Tan, Liu, Liu, Kaufman, and
  Zhou}]{xie2024progressive}
Xie, D.; Xu, Z.; Hong, Y.; Tan, H.; Liu, D.; Liu, F.; Kaufman, A.; and Zhou, Y.
  2024.
\newblock Progressive autoregressive video diffusion models.
\newblock \emph{arXiv preprint arXiv:2410.08151}.

\bibitem[{Xue et~al.(2022)Xue, Hang, Zeng, Sun, Liu, Yang, Fu, and
  Guo}]{xue2022advancing}
Xue, H.; Hang, T.; Zeng, Y.; Sun, Y.; Liu, B.; Yang, H.; Fu, J.; and Guo, B.
  2022.
\newblock Advancing high-resolution video-language representation with
  large-scale video transcriptions.
\newblock In \emph{Proceedings of the IEEE/CVF Conference on Computer Vision
  and Pattern Recognition}, 5036--5045.

\bibitem[{Yan et~al.(2024)Yan, Cai, Wang, Zhou, Huang, and Yang}]{yan2024long}
Yan, X.; Cai, Y.; Wang, Q.; Zhou, Y.; Huang, W.; and Yang, H. 2024.
\newblock Long Video Diffusion Generation with Segmented Cross-Attention and
  Content-Rich Video Data Curation.
\newblock \emph{arXiv preprint arXiv:2412.01316}.

\bibitem[{Yang et~al.(2024)Yang, Teng, Zheng, Ding, Huang, Xu, Yang, Hong,
  Zhang, Feng et~al.}]{yang2024cogvideox}
Yang, Z.; Teng, J.; Zheng, W.; Ding, M.; Huang, S.; Xu, J.; Yang, Y.; Hong, W.;
  Zhang, X.; Feng, G.; et~al. 2024.
\newblock Cogvideox: Text-to-video diffusion models with an expert transformer.
\newblock \emph{arXiv preprint arXiv:2408.06072}.

\bibitem[{Yin et~al.(2023)Yin, Wu, Yang, Wang, Wang, Ni, Yang, Li, Liu, Yang
  et~al.}]{yin2023nuwa}
Yin, S.; Wu, C.; Yang, H.; Wang, J.; Wang, X.; Ni, M.; Yang, Z.; Li, L.; Liu,
  S.; Yang, F.; et~al. 2023.
\newblock NUWA-XL: Diffusion over Diffusion for eXtremely Long Video
  Generation.
\newblock In \emph{The 61st Annual Meeting Of The Association For Computational
  Linguistics}.

\bibitem[{Yin et~al.(2024)Yin, Zhang, Zhang, Freeman, Durand, Shechtman, and
  Huang}]{yin2024slow}
Yin, T.; Zhang, Q.; Zhang, R.; Freeman, W.~T.; Durand, F.; Shechtman, E.; and
  Huang, X. 2024.
\newblock From slow bidirectional to fast autoregressive video diffusion
  models.
\newblock \emph{arXiv preprint arXiv:2412.07772}, 2.

\bibitem[{Yu et~al.(2025)Yu, Hahn, Kondratyuk, Shin, Gupta, Lezama, Essa, Ross,
  and Huang}]{yu2025malt}
Yu, S.; Hahn, M.; Kondratyuk, D.; Shin, J.; Gupta, A.; Lezama, J.; Essa, I.;
  Ross, D.; and Huang, J. 2025.
\newblock MALT Diffusion: Memory-Augmented Latent Transformers for Any-Length
  Video Generation.
\newblock \emph{arXiv preprint arXiv:2502.12632}.

\bibitem[{Zhang and Agrawala(2025)}]{zhang2025packing}
Zhang, L.; and Agrawala, M. 2025.
\newblock Packing input frame context in next-frame prediction models for video
  generation.
\newblock \emph{arXiv preprint arXiv:2504.12626}.

\bibitem[{Zhang et~al.(2025{\natexlab{a}})Zhang, Bi, Hong, Zhang, Luan, Yang,
  Sunkavalli, Freeman, and Tan}]{zhang2025test}
Zhang, T.; Bi, S.; Hong, Y.; Zhang, K.; Luan, F.; Yang, S.; Sunkavalli, K.;
  Freeman, W.~T.; and Tan, H. 2025{\natexlab{a}}.
\newblock Test-time training done right.
\newblock \emph{arXiv preprint arXiv:2505.23884}.

\bibitem[{Zhang et~al.(2025{\natexlab{b}})Zhang, Jiang, Ma, Lu, Huang, Yuan,
  and Duan}]{zhang2025generative}
Zhang, Y.; Jiang, J.; Ma, G.; Lu, Z.; Huang, H.; Yuan, J.; and Duan, N.
  2025{\natexlab{b}}.
\newblock Generative pre-trained autoregressive diffusion transformer.
\newblock \emph{arXiv preprint arXiv:2505.07344}.

\bibitem[{Zhang et~al.(2025{\natexlab{c}})Zhang, Xing, Xia, Liu, Peng, Tao,
  Wan, Lo, and Jia}]{zhang2025training}
Zhang, Y.; Xing, J.; Xia, B.; Liu, S.; Peng, B.; Tao, X.; Wan, P.; Lo, E.; and
  Jia, J. 2025{\natexlab{c}}.
\newblock Training-Free Efficient Video Generation via Dynamic Token Carving.
\newblock \emph{arXiv preprint arXiv:2505.16864}.

\bibitem[{Zhao et~al.(2025)Zhao, He, Chen, Zhu, Li, and Zhu}]{zhao2025riflex}
Zhao, M.; He, G.; Chen, Y.; Zhu, H.; Li, C.; and Zhu, J. 2025.
\newblock Riflex: A free lunch for length extrapolation in video diffusion
  transformers.
\newblock \emph{arXiv preprint arXiv:2502.15894}.

\bibitem[{Zheng et~al.(2025)Zheng, Yuan, Wang, Huang, Ma, and
  Duan}]{zheng2025frame}
Zheng, G.; Yuan, J.; Wang, B.; Huang, H.; Ma, G.; and Duan, N. 2025.
\newblock Frame-Level Captions for Long Video Generation with Complex Multi
  Scenes.
\newblock \emph{arXiv preprint arXiv:2505.20827}.

\bibitem[{Zheng et~al.(2024{\natexlab{a}})Zheng, Xu, Huang, Ma, Liu, Shu, Pang,
  Tang, Chen, Yang et~al.}]{zheng2024videogen}
Zheng, M.; Xu, Y.; Huang, H.; Ma, X.; Liu, Y.; Shu, W.; Pang, Y.; Tang, F.;
  Chen, Q.; Yang, H.; et~al. 2024{\natexlab{a}}.
\newblock VideoGen-of-Thought: A Collaborative Framework for Multi-Shot Video
  Generation.
\newblock \emph{arXiv preprint arXiv:2412.02259}.

\bibitem[{Zheng et~al.(2024{\natexlab{b}})Zheng, Peng, Yang, Shen, Li, Liu,
  Zhou, Li, and You}]{zheng2024open}
Zheng, Z.; Peng, X.; Yang, T.; Shen, C.; Li, S.; Liu, H.; Zhou, Y.; Li, T.; and
  You, Y. 2024{\natexlab{b}}.
\newblock Open-sora: Democratizing efficient video production for all.
\newblock \emph{arXiv preprint arXiv:2412.20404}.

\bibitem[{Zhou et~al.(2024{\natexlab{a}})Zhou, Wang, Cai, and
  Yang}]{zhou2024allegro}
Zhou, Y.; Wang, Q.; Cai, Y.; and Yang, H. 2024{\natexlab{a}}.
\newblock Allegro: Open the black box of commercial-level video generation
  model.
\newblock \emph{arXiv preprint arXiv:2410.15458}.

\bibitem[{Zhou et~al.(2024{\natexlab{b}})Zhou, Zhou, Cheng, Feng, and
  Hou}]{zhou2024storydiffusion}
Zhou, Y.; Zhou, D.; Cheng, M.-M.; Feng, J.; and Hou, Q. 2024{\natexlab{b}}.
\newblock Storydiffusion: Consistent self-attention for long-range image and
  video generation.
\newblock \emph{Advances in Neural Information Processing Systems}, 37:
  110315--110340.

\bibitem[{Zhu et~al.(2023)Zhu, Yang, He, Wang, Tuo, Cheng, Gao, Song, and
  Fu}]{zhu2023moviefactory}
Zhu, J.; Yang, H.; He, H.; Wang, W.; Tuo, Z.; Cheng, W.-H.; Gao, L.; Song, J.;
  and Fu, J. 2023.
\newblock Moviefactory: Automatic movie creation from text using large
  generative models for language and images.
\newblock In \emph{Proceedings of the 31st ACM International Conference on
  Multimedia}, 9313--9319.

\bibitem[{Zhuang et~al.(2025)Zhuang, Huang, Zhang, Wang, Fu, Yang, Sun, Li, and
  Wang}]{zhuang2025video}
Zhuang, S.; Huang, Z.; Zhang, Y.; Wang, F.; Fu, C.; Yang, B.; Sun, C.; Li, C.;
  and Wang, Y. 2025.
\newblock Video-GPT via Next Clip Diffusion.
\newblock \emph{arXiv preprint arXiv:2505.12489}.

\bibitem[{Zhuang et~al.(2024)Zhuang, Li, Chen, Wang, Liu, Qiao, and
  Wang}]{zhuang2024vlogger}
Zhuang, S.; Li, K.; Chen, X.; Wang, Y.; Liu, Z.; Qiao, Y.; and Wang, Y. 2024.
\newblock Vlogger: Make your dream a vlog.
\newblock In \emph{Proceedings of the IEEE/CVF Conference on Computer Vision
  and Pattern Recognition}, 8806--8817.

\end{thebibliography}

\end{document}